\documentclass[acmsmall]{acmart}

\usepackage{multirow}
\usepackage{makecell}
\usepackage[caption=false,font=footnotesize]{subfig}
\usepackage[linesnumbered,ruled,vlined]{algorithm2e}
\usepackage[flushleft]{threeparttable}

\AtBeginDocument{%
  }

\setcopyright{acmlicensed}
\copyrightyear{2024}
\acmYear{2024}
\acmDOI{xxxx/xxxx}

\acmJournal{TOMM}
\acmVolume{x}
\acmNumber{x}
\acmArticle{x}
\acmMonth{5}

\begin{document}

\title{A Multi-task Adversarial Attack Against Face Authentication}

\author{Hanrui Wang}
\email{hanrui_wang@nii.ac.jp}
\orcid{0009-0005-0498-2712}
\affiliation{%
  \institution{National Institute of Informatics}
  \city{Tokyo}
  \country{Japan}
}

\author{Shuo Wang}
\email{wangshuosj@sjtu.edu.cn}
\orcid{0000-0001-8938-2364}
\affiliation{%
  \institution{Shanghai Jiao Tong University}
  \city{Shanghai}
  \country{China}
}

\author{Cunjian Chen}
\email{cunjian.chen@monash.edu}
\orcid{0000-0002-2926-9762}
\affiliation{%
  \institution{Monash University}
  \city{Clayton}
  \state{Victoria}
  \country{Australia}
}

\author{Massimo Tistarelli}
\email{tista@uniss.it}
\orcid{0000-0002-3406-3048}
\affiliation{%
  \institution{University of Sassari}
  \city{Sassari}
  \country{Italy}
}

\author{Zhe Jin}
\authornote{Corresponding Author}
\email{jinzhe@ahu.edu.cn}
\orcid{0000-0003-4501-7992}
\affiliation{%
  \institution{Anhui University}
  \city{Hefei}
  \state{Anhui}
  \country{China}
}


\begin{abstract}
Deep-learning-based identity management systems, such as face authentication systems, are vulnerable to adversarial attacks. However, existing attacks are typically designed for single-task purposes, which means they are tailored to exploit vulnerabilities unique to the individual target rather than being adaptable for multiple users or systems. This limitation makes them unsuitable for certain attack scenarios, such as morphing, universal, transferable, and counter attacks. In this paper, we propose a multi-task adversarial attack algorithm called MTADV that are adaptable for multiple users or systems. By interpreting these scenarios as multi-task attacks, MTADV is applicable to both single- and multi-task attacks, and feasible in the white- and gray-box settings. Furthermore, MTADV is effective against various face datasets, including LFW, CelebA, and CelebA-HQ, and can work with different deep learning models, such as FaceNet, InsightFace, and CurricularFace. Importantly, MTADV retains its feasibility as a single-task attack targeting a single user/system. To the best of our knowledge, MTADV is the first adversarial attack method that can target all of the aforementioned scenarios in one algorithm.
\end{abstract}



\received{30 March 2023}
\received[revised]{6 December 2023}
\received[accepted]{12 May 2024}

\maketitle

\section{Introduction}
\label{Introduction}
\begin{figure}[!htb]
	\setlength{\abovecaptionskip}{0.1cm}
	\setlength{\belowcaptionskip}{0.1cm}
    \centering
        \includegraphics[width=4in]{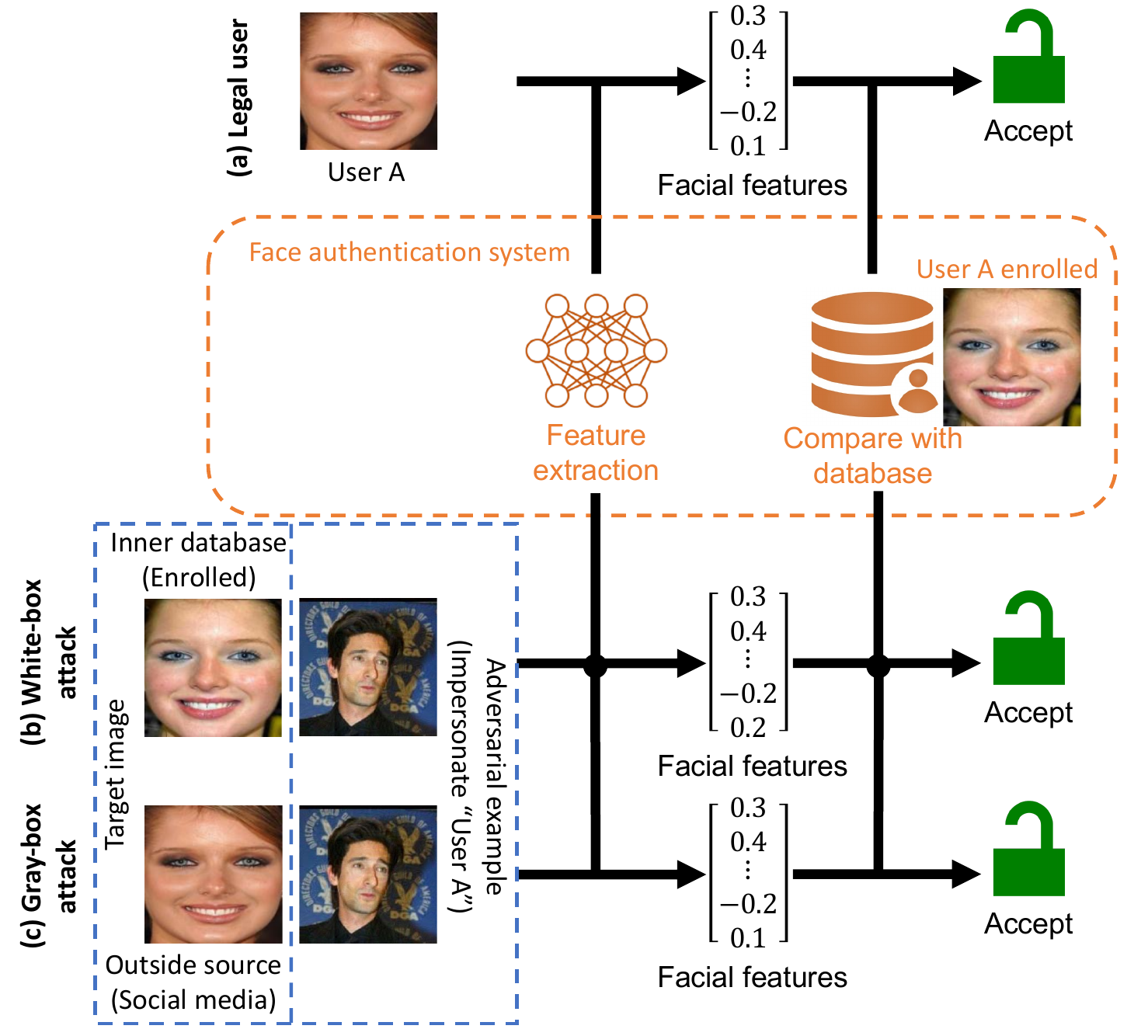}
        \Description{Adversarial attacks against face authentication}
        \caption{The single-task adversarial attacks against face authentication. (a) A legal user accesses the system. (b) It is a white-box attack when the target image is identical to that enrolled in the database, i.e., the database is known to the attacker. (c) It is a gray-box attack when the target image is different from that enrolled, i.e., the database is unknown.}
        \label{fig_white-gray}
\end{figure}

In recent years, deep learning technologies have achieved remarkable performance in security applications such as face authentication. However, security concerns posed by adversarial attacks remain as these attacks uncover the vulnerabilities of deep learning models. An adversarial attack launched against a face authentication system can be depicted as a process to mislead the decision from rejection to acceptance by feeding a deceptive input, also known as an adversarial example \cite{Kur16}. The threat models revealed by existing adversarial attacks can be categorised into three types: 1) white box -  an attack with the knowledge of both the feature extractor and the target image from the database, which indicates the deep learning model and database are known to the attacker, as shown in Fig. \ref{fig_white-gray}(b); 2) gray box - an attack with the knowledge of the feature extractor, but without the knowledge of the database, as shown in Fig. \ref{fig_white-gray}(c); 3) black box - an attack without the knowledge of the feature extractor, but the target image must be identical with the one enrolled in the database, which is claimed as requiring the knowledge of the database \cite{wang2021fg}. Note that such perceived attacks are only intended for attacking authentication systems instead of classification systems.

\begin{figure}[!htb]
	\setlength{\abovecaptionskip}{0.1cm}
	\setlength{\belowcaptionskip}{0.1cm}
    \centering
        \includegraphics[width=3.5in]{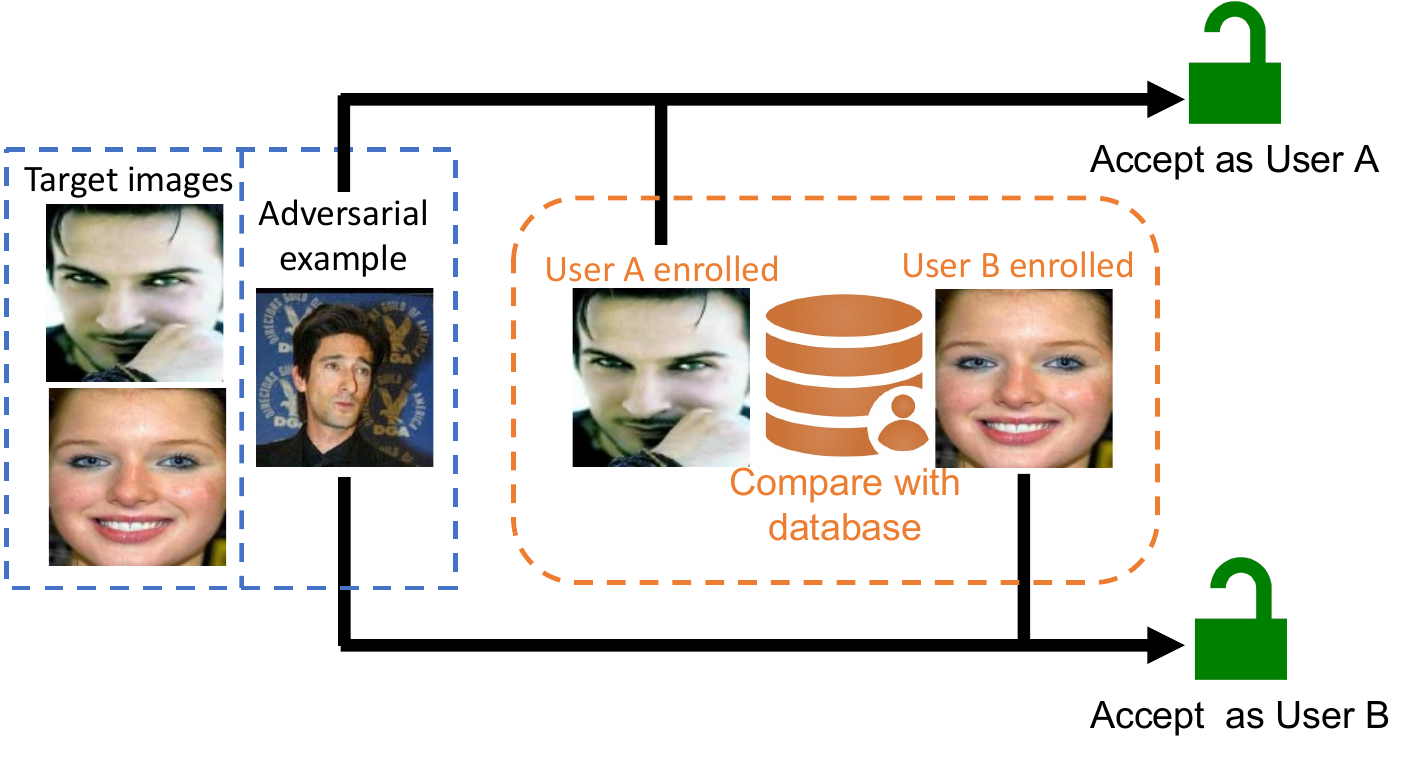}
        \Description{The multi-task adversarial attack impersonating multiple users}
        \caption{The multi-task adversarial attack impersonating multiple users. A single adversarial example can gain illegal access acting as multiple target users.}
        \label{fig_multi-user}
\end{figure}

\begin{figure}[!htb]
	\setlength{\abovecaptionskip}{0.1cm}
	\setlength{\belowcaptionskip}{0.1cm}
    \centering
        \includegraphics[width=\linewidth]{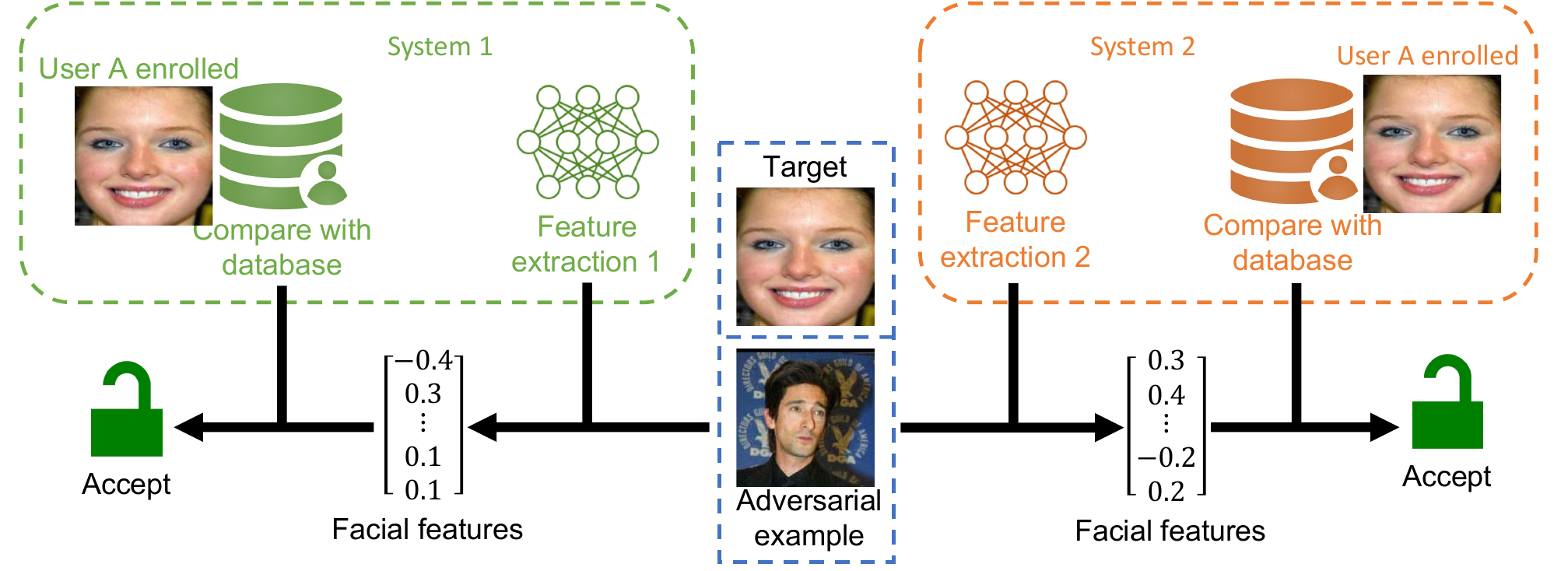}
        \Description{The multi-task adversarial attack attacking multiple systems}
        \caption{The multi-task adversarial attack attacking multiple systems. A single adversarial example can gain illegal access to multiple systems where the identical target user enrolls.}
        \label{fig_multi-system}
\end{figure}

Despite the variety of threat models, current adversarial attacks primarily focus on \textbf{single-task} objectives that one adversarial example can only impersonate one specific user within a given system. However, in some cases, an attacker may want to use a single input to impersonate multiple legitimate users or gain unauthorized access to various systems \cite{ferrara2014magic}, which we interpret them as \textbf{multi-task} attacks. Unfortunately, existing single-task adversarial attacks are not suitable for these more complex multi-task attack scenarios. In this paper, we investigate four representative multi-task attack scenarios that can reflect real world cases:
\begin{itemize}
    \item \textit{Morphing attack (MA): Multiple subjects can pass the auto-gate at the Customs using the same passport.} This is the most fundamental attack scenario for impersonating multiple users, as illustrated in Fig. \ref{fig_multi-user}. 
    \item \textit{Universal attack (UA): The input is an average face, which can match numerous human beings.} This is a more powerful impersonation attack. In our context, the attack does not require any target image.
    \item \textit{Transferable attack (TA): A single input can gain illegal access to multiple systems where the identical target user enrolls.} This is a basic attack scenario for attacking multiple systems, as illustrated in Fig. \ref{fig_multi-system}.
    \item \textit{Counterattack (CA): The attack can defeat adversarial defenses.} This is a special application of multi-task attacks when assuming the defense as another system apart from the feature extractor.
\end{itemize}

While the multi-task attack scenarios mentioned above are relevant in real-world situations, there have been limited endeavors in this area, and the available research primarily addresses only a single scenario (either MA or TA) \cite{haleta2021multitask,ghamizi2022adversarial,liu2023attention}. Therefore, we introduce MTADV, an abbreviation for ``multi-task adversarial attack,'' to address this deficiency. Additionally, since single-task attacks (STs) represent the fundamental case in adversarial attacks, MTADV retains its applicability as a single-task attack, ensuring its effectiveness against face authentication threat models. To utilise MTADV, we present a novel objective function that seeks an adversarial example positioned at the midpoint among all target images in the feature space. Leveraging this proposed objective function, MTADV can be employed for single- and multi-task attacks (ST, MA, UA, TA, and CA), both in white-box and gray-box settings, across various face datasets (labeled faces in the wild (LFW), CelebA, and high-quality CelebA (CelebA-HQ)), and with different deep learning models (FaceNet, InsightFace, and CurricularFace). To the best of our knowledge, MTADV stands out as the first adversarial attack capable of addressing all the aforementioned scenarios through a single algorithm. Furthermore, we conduct comprehensive ablation studies to illustrate the effects of various parameters on the attack success rate, efficiency, and image quality. Our code is released at \href{https://github.com/azrealwang/mtadv}{https://github.com/azrealwang/mtadv}.
Our contributions are summarized as follows:
\begin{itemize}
    \item We firstly proposed a multi-task adversarial attack called MTADV, which is capable of targeting multiple users or systems with a single attempt. This adaptable approach offers a more versatile solution for handling complex attack scenarios that involve multiple targets simultaneously.
    \item We further propose a novel objective function, which enables MTADV suited for a wide variety of attack scenarios by a single algorithm, including single-/multi-task scenarios and white-/gray-box settings.
    \item We experimentally demonstrate MTADV's effectiveness to five representative real world's attack scenarios (four multi-task and one single-task) and its generalization against three face datasets and deep learning models.
    \item We conduct thorough ablation studies, then provide a detailed analysis of the correlation between various parameters and the attack performance.
\end{itemize}
\section{Background}
\label{Literature}
In this section, we will provide a concise overview of the foundations of face authentication systems and different formulations of adversarial attacks. Note that we will focus on these foundational aspects within the context of targeted/impersonation attacks, as our primary aim is to assess the security of authentication systems.

\subsection{Deep learning model for face authentication}
Deep-learning-based face authentication has achieved superior performance. The training loss functions \cite{Sch15,liu2017sphereface,wang2018cosface,Den19} and optimized network architectures \cite{Tai14,Sun15,liu2017sphereface,Den19}, in particular, play significant roles in gaining satisfactory accuracy. For instance, FaceNet \cite{Sch15} introduces the triplet loss to increase the Euclidean margin between identities in the feature space. CosFace \cite{wang2018cosface} adopts the significant margin cosine loss to maximize the cosine margin. SphereNet \cite{liu2017sphereface} adopts the angular softmax loss in a 64-layer ResNet \cite{HeK16} network to learn angularly discriminative features. InsightFace (ArcFace) \cite{Den19} combines the ResNet \cite{HeK16} and squeeze-and-excitation network (SENet) \cite{hu2018squeeze} with an improve residual block \cite{Den19}, supervised by an additive angular margin loss. CurricularFace \cite{huang2020curricularface}, one of the best and recently proposed deep learning models for face authentication in terms of accuracy, adopts an adaptive curriculum learning loss to achieve a novel training strategy for face authentication, which uses easy samples (front and clear samples) in the early training stage and hard ones in the later stage, and then adaptively adjusts the relative importance of easy and hard samples during different training stages according to their corresponding difficulties.

\subsection{Adversarial attack for classification}
\label{adversaral_attacks}
The fast gradient sign method (FGSM) \cite{Goo14}, projected gradient descent (PGD) \cite{madry2018towards}, DeepFool \cite{Moo16}, and the Carlini \& Wagner attack (CW) \cite{Car171,Car17} represent standard benchmarks for adversarial attacks. While these techniques were originally developed for classification, they have now found widespread application in diverse tasks, such as face recognition, for the purpose of comparison. Specifically, FGSM \cite{Goo14} was proposed by Goodfellow et al. to instantly produce adversarial examples to meet the demands for generating a large number of adversarial examples for training and authentication. DeepFool \cite{Moo16} was proposed by Moosavi-Dezfooli et al., which introduces slight perturbations to the image to attack deep learning models. It revealed that deep learning models are not robust even with slight perturbations. PGD \cite{madry2018towards} was proposed as an iterative version of the one-step attack FGSM to find the most adversarial examples with acceptable efficiency. CW \cite{Car171} was the first approach designed to counterattack the defenses of \cite{szegedy2014intriguing} and \cite{Goo14}. Although it is computationally expensive, CW is regarded more aggressive than other attacks. Nonetheless, all of the aforementioned methods are impractical for tasks like face authentication, especially in the gray-box setting, or for handling multi-task attack scenarios.

\subsection{Adversarial attack for face authentication}
Apart from above general methods, there are various adversarial attacks specifically tailored to target face authentication systems. Dabouei et al. \cite{dabouei2019fast} proposed an efficient algorithm directly manipulating the landmarks of the face image to produce a geometrically perturbed adversarial example. Yang et al. \cite{Yan21} proposed a generative-adversarial-network-based (GAN-based) adversarial attack in the gray-box setting, namely \(A^{3}GN\). However, \(A^{3}GN\) requires at least five target images from each target user for training and inference. Wang et al. \cite{wang2021fg} introduced a similarity-based gray-box attack, which has demonstrated the best performance in both single-task white-box and gray-box scenarios. Furthermore, there is a growing interest in the development of black-box adversarial attacks. Deb et al. \cite{Deb19} proposed a GAN-based black-box method, namely AdvFaces, to generate more natural face images. Zhong et al. \cite{zhong2020towards} developed the deep feature aggregation network, a model trained to generate transferable adversarial examples without requiring knowledge of the target model. Li et al. \cite{li2023sibling} presented the sibling-attack, which utilizes multi-task learning techniques to extract information from various tasks for a single target. However, it's worth noting that all of these methods are not suitable for multi-task attack scenarios. The sibling-attack, for instance, incorporates multi-task learning to target a single entity, while our approach can target multiple entities simultaneously. Additionally, in single-task attack scenarios, these methods do not outperform our MTADV in our threat models, which establishes MTADV as the leading attack method. Furthermore, none of these attacks effectively addresses the challenge of gray-box attack scenarios when the enrolled image differs from the target image.

\subsection{Adversarial attack for multiple tasks}
Very few published adversarial attacks are designed for multi-task scenarios, and none of the existing works are suitable for all scenarios (MA, UA, TA, and CA) or achieve comparable performance as our MTADV. Specifically, Haleta et al. \cite{haleta2021multitask} proposed a multi-task adversarial attack, but it's only applicable for transferable attacks against classification systems. Furthermore, their image quality is considerably lower (using \(\epsilon=16/255\)) compared to our approach (using \(\epsilon=8/255\)) to attain the claimed performance. Ghamizi et al. \cite{ghamizi2022adversarial} and Liu et al. \cite{liu2023attention} both introduced multi-task adversarial attacks aimed at multi-task learning models. However, their applicable attack scenario is limited to morphing attacks (targeting multiple targets) because multi-task learning models typically involve a restricted set of task types, unlike our MTADV, which can be applied to as many as 50 targets/tasks simultaneously.


\section{Multi-task adversarial attack}
\label{design}
This section presents the proposed MTADV. Section \ref{adv_model} introduces the threat models of MTADV. Section \ref{problem} addresses the problem of the adversarial attack against face authentication. Section \ref{loss} theoretically explains the feasibility of the proposed objective function, and the algorithm of MTADV is presented in Section \ref{Algorithm}.

\subsection{Threat model}
\label{adv_model}
A face authentication system involves two key components, including a deep learning model as the feature extractor and a set of facial features stored in a database (as illustrated in Fig. \ref{fig_white-gray}(a)). To enhance the security, some authentication systems may integrate a defense module against adversarial attacks. An adversarial attack for impersonation may require the knowledge of one or multiple aforementioned items and a target image, which is the threat model of the attack. Table \ref{tab_threat_model} lists the threat models of MTADV in the investigated single-/multi-task attack scenarios. Compared MTADV with traditional white-/gray-/black-box settings, MTADV is a more powerful gray-box attack, but it is not applicable to the black-box setting. However, MTADV is the only method capable to attack without requiring the target image and is also applicable to the system with the defense module.
\begin{table}[!htb]
	\setlength{\abovecaptionskip}{0.1cm}
	\setlength{\belowcaptionskip}{0.1cm}
    \caption{Threat models of MTADV}
    \label{tab_threat_model}
    \centering
    \begin{threeparttable}
    \setlength{\tabcolsep}{3mm}{\begin{tabular}{ccccc}
        \hline
        Threat model&Deep learning model&Database&Defense&Target image\\
        \hline
        \hline
        White box&Compulsory&Compulsory&Not Applicable&Compulsory\\
        Gray box&Compulsory&Optional&Not Applicable&Compulsory\\
        Black box$^*$&Optional&Compulsory&Not Applicable&Compulsory\\
        \hline
        MTADV-ST&Compulsory&Optional&Not Applicable&Compulsory\\
        MTADV-MA&Compulsory&Optional&Not Applicable&Compulsory\\
        MTADV-UA&Compulsory&Optional&Not Applicable&Optional\\
        MTADV-TA&Compulsory&Optional&Not Applicable&Compulsory\\
        MTADV-CA&Compulsory&Optional&Compulsory&Compulsory\\
        \hline
    \end{tabular}}
    \begin{tablenotes}
          \footnotesize
          \item $^{*}$ Compared between the top and bottom panels, MTADV is a more powerful gray-box attack, but it is not applicable to the black-box setting.
    \end{tablenotes}
    \end{threeparttable}
\end{table}


Furthermore, the practice for attackers to steal the information from each component is discussed as follows. First of all, for any system, the three system components, i.e., the deep learning model, database, and defense if existing, can be deployed in the same server or divided into different servers. It is highly possible that the attacker need individual access to these servers. Therefore, as the database must be deployed in the production server and only accessible by system engineers and database administrators, it is most difficult to steal. In contrast, the deep learning model and defense module are partial system code (implementation), which is also public to the development team (e.g., outsourcing team). In addition, if the implementation is based on commercial application programming interfaces (APIs) (e.g., Amazon Rekognition) or deployed in a cloud server (e.g., Google Cloud Platform), the suppliers may also have the full knowledge of the specific modules even the token \cite{lai2021efficient}. All these exposed information can be potential threats. However, when comparing between the deep learning model and defense module, defense module is less likely to be stole because there are numerous public pretrained models and commercial APIs for face authentication, but there is no commonly agreement on the defense strategy, which means the defense module may be more customised. Lastly, the target image, which is a compulsory requirement by most adversarial impersonation attacks, is easiest to gain unless it have to be the identical one from the database like in the white-box and black-box settings. For the gray-box setting, the target image can be any face image from the target user. In summary, the difficulties of gaining the knowledge for the attackers should be ordered as:
Database$>$Defense$>$Deep learning model$>$Target image. In this order, MTADV may be applicable to more general usage.

\subsection{Problem definition}
\label{problem}
Our goal of attacking the authentication system is to gain illegal access to the system, as shown in Figs. \ref{fig_white-gray}-\ref{fig_multi-system}. Let \(X^{source}\) be the source face image of the adversary, which will be perturbed to produce the adversarial example \(X^{adv}\). Specifically, define \(X^{adv}\) as \(X^{source}\) with perturbations \(\delta\), so
\begin{equation}
    X^{adv}=X^{source}+\delta,\ s.t.\ ||\delta||_\infty<\epsilon,
\end{equation}
where $\epsilon$ is the pre-defined perturbation size and is small enough to evade human inspection. Let \(X^{enrolled}\) represent the enrolled face image of a legal user \(X\) (the target user). Thus, \(X^{enrolled}\) is stored in the database and unknown to the adversary in the gray-box setting. Let \(f\left(\cdot \right)\) denote the function of the feature extractor in the face authentication system, whose output is the feature vector (facial features) for comparison. Let \(||\cdot||\in[0,1]\) denote the L2-norm distance, which equals the dissimilarity ranging from 0 to 1 (a lower value indicates a higher similarity). Therefore, an adversarial attack can be regarded as successful when:
\begin{equation}
    ||f(X^{adv})-f(X^{enrolled})||\leq\tau,
    \label{eq_problem}
\end{equation}
where $\tau$ is a pre-defined dissimilarity threshold.

\subsection{Objective function}
\label{loss}
MTADV is designed for multi-task attacks, i.e., MA, UA, TA, and CA. Likewise, MTADV preserves its feasibility in the basic ST scenario, which is regarded as a special case in MTADV when targeting a single system and target. In single-task attacks, the adversarial example is learned until it is highly similar to the target image in the feature space, as indicated in Eq. \eqref{eq_loss_st}. Such a deceptive adversarial example can successfully breach the authentication system by comparing it with the target or enrolled image.
\begin{equation}
    J_{ST}(X^{adv})=||f(X^{adv})-f(X)||.
    \label{eq_loss_st}
\end{equation}

Nonetheless, in multi-task attacks, a successful attack necessitates that the distance in the feature space between the single adversarial example and each target must be smaller than a predefined threshold. However, this results in that the adversarial example cannot be extremely close to either target in the feature space. Therefore, the attack performance of multi-task attacks will drop compared with that of single-task attacks.

To achieve the above objectives, a new objective function is proposed as follows:
\begin{equation}
    \begin{aligned}
        J_{MT}(X^{adv})=\Sigma_{m=1}^{M}(\frac{\Sigma_{n=1}^{N}||f_m(X^{adv})-f_m(X_n)||}{\tau_m}),
    \end{aligned}
    \label{eq_loss_mtadv}
\end{equation}
where $\tau_m$ is the pre-defined dissimilarity threshold of $m$-th face authentication system $f_m(\cdot)$ and $X_n$ is the target face image of the $n$-th target user. $M$ and $N$ indicate that there are $M$ target face authentication systems and $N$ target users. $\tau_m$ normalizes the impact of different systems, so only compulsory for attacking multiple systems. For attacking multiple users, $\tau_m$ are all same, so Eq. \eqref{eq_loss_mtadv} can be simplified as:
\begin{equation}
    J_{MT}(X^{adv})=\frac{\Sigma_{n=1}^{N}||f(X^{adv})-f(X_n)||}{N}.
\end{equation}
The single-task attack is a special case only when \(M\ (systems)=N\ (targets)=1\).

\subsection{Geometric proof}
\label{proof}
\begin{figure}[!htb]
	\setlength{\abovecaptionskip}{0.1cm}
	\setlength{\belowcaptionskip}{0.1cm}
	\centering
		\includegraphics[width=\linewidth]{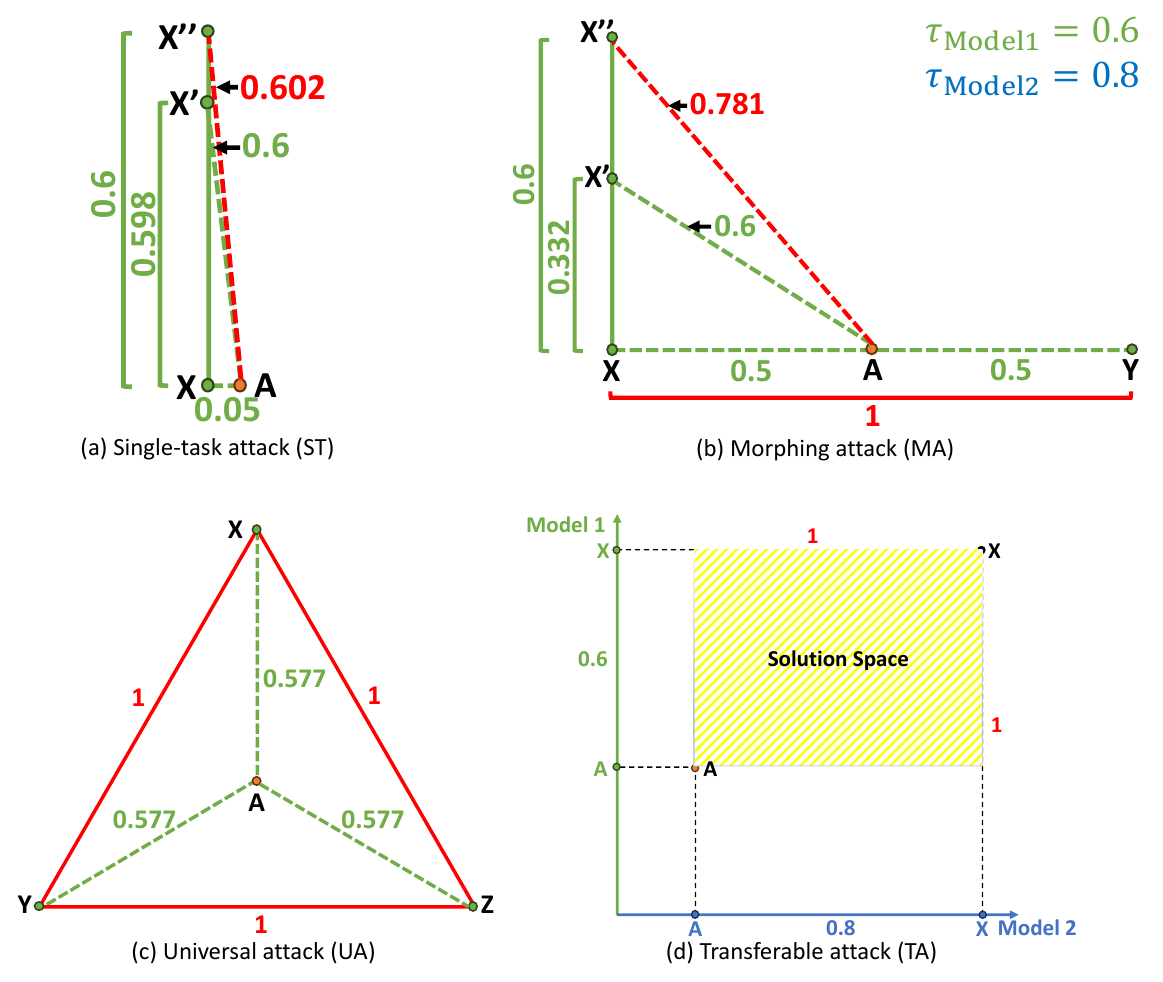}
        \Description{Geometric proof of the feasibility of MTADV in various attack scenarios}
		\caption{Geometric proof of the feasibility of MTADV in various attack scenarios. $X$, $Y$, and $Z$ denote targets. $X'$ and $X''$ denote other face images of $X$, regarding to the gray-box attack scenarios. $A$ denotes the adversarial example. The numbers in green represent the distances smaller than the threshold of Model 1 (FaceNet), whose threshold is approximate 0.6. The numbers in red represent the distances higher than the threshold, which indicates they are different subjects. The information in blue belongs to another model (InsightFace), whose threshold is approximate 0.8, regarding to the transferable attack scenario. These numbers are obtained based on our worst assumptions and geometric calculations.}
		\label{fig_geometry}
\end{figure}

In this section, we theoretically analyze MTADV's feasibility and performance using geometric principles. These analyses showcase the simplicity and effectiveness of our methodology while illustrating that it has achieved the theoretical maximum performance within the threat models we consider. First, we define the distance between two images in the feature space ranging from 0 (same) to 1. Hence, any two images are regarded as the same subject when \(||f(X)-f(X')||\leq\tau\), where $\tau$ is the distance threshold and $X/X^*$ belong to the same subject. In contrast, they are different subjects when \(||f(X)-f(Y)||\in(\tau,1]\), where $X,Y,...$ represent different subjects. Second, we assume the worst case for different subjects is \(||f(X)-f(Y)||=1\). Therefore, the real experimental performance will be always better than the theoretical calculation. Finally, the analyses are conducted in two models (FaceNet in green and InsightFace in blue) to simulate all five attack scenarios (ST, MA, UA, TA, and CA).

\textbf{Single-task attack (MTADV-ST)}:
As shown in Figure \ref{fig_geometry}(a), contributed by our loss function (Eq. \eqref{eq_loss_mtadv}), the distance between the target image $X$ and adversarial example $A$ can be very small, represented by $\ell\approx 0$. Therefore, the white-box attack performance is reasonable to achieve 100\% as it is estimated via:
\begin{equation}
    D_{ST-W}=\ell\ll\tau.
\end{equation}
However, in the gray-box attack, the distance is computed via:
\begin{equation}
    D_{ST-G}=\sqrt{||f(X)-f(X')||^2+\ell^2}, s.t.\ ||f(X)-f(X')||\leq\tau\ and\ \ell\approx 0.
    \label{eq_d_st_g}
\end{equation}
Eq. \eqref{eq_d_st_g} indicates that it is highly possible that $D_{ST-G}$ is smaller than $\tau$, which reflects the performance in Table \ref{tab_ST}. Note that as $\ell$ is sufficiently small, the results in Table \ref{tab_ST} have demonstrated that the error of gray-box attacks is caused by the system error, which cannot be further improved in theory.

\textbf{Morphing attack (MTADV-MA)}:
As shown in Figure \ref{fig_geometry}(b), the best adversarial example of MTADV-MA should obtain the same distance from either of the two targets. It results in the distance under the white-box attack setting which can be estimated via:
\begin{equation}
    D_{MA-W}\in(\frac{\tau}{2},0.5].
\end{equation}
As the majority of thresholds of existing face recognition models are higher than 0.5, MTADV-MA can achieve 100\% in the white-box setting. However, as \(D_{MA-W}>D_{ST-W}\), the performance of MTADV-MA in the gray-box setting will be lower than that of MTADV-ST:
\begin{equation}
    D_{MA-G}=\sqrt{||f(X)-f(X')||^2+D_{MA-W}^2}>D_{ST-G}
\end{equation}

\textbf{Universal attack (MTADV-UA)}:
As explained in Figure \ref{fig_geometry}(b), MTADV-MA computes the adversarial example in the middle of two target points. Likewise, if there are more targets, which is defined as MTADV-UA, the adversarial example should be the middle point of all target points. For example, as shown in Figure \ref{fig_geometry}(c), to attack three targets, the center point (adversarial example $A$) is computed from two points to three points. In our worst assumption, the distance between $A$ and each target is still smaller than the threshold, which indicates the success of the universal attack. However, the geometric result implies that the performance of MTADV-UA will be largely reduced compared to MTADV-MA as the distance between targets and adversarial example increases, and this performance will be further decreased if more targets included. In addition, Figure \ref{fig_geometry}(c) indicates that attacking a model with a higher distance threshold will be easier as the tolerance is enlarged.

\textbf{Transferable attack (MTADV-TA)}:
We assume that the two models are non-correlated. The adversarial example and target will be compared only in their individual feature space after mapping. As shown in Figure \ref{fig_geometry}(d), there is a considerable large solution space for MTADV-TA. Moreover, compared Figure \ref{fig_geometry}(d) with Figure \ref{fig_geometry}(b), it indicates that the performance of MTADV-TA can be higher than MTADV-MA as $A$ can be more closer to $X$ in the individual feature space, which reflects the results in Tables \ref{tab_MA} and \ref{tab_TA}.

\textbf{Counterattack (MTADV-CA)}:
MTADV-CA is an application of MTADV-TA specially against defenses by regarding the defense modules as other models. Therefore, the feasibility of MTADV-TA applies to MTADV-CA.

Lastly, as observed in the analyses above, it becomes increasingly challenging to execute successful attacks as the number of targets (either subjects or systems) grows. This complexity arises from the need to solve intricate geometric problems. Therefore, we believe that a more comprehensive mathematical formulation could potentially enhance performance. This aspect will be a focus of our future investigations.

\subsection{Algorithm}
\label{Algorithm}
\begin{algorithm}[!b]
    \DontPrintSemicolon
    \caption{MTADV}
    \label{alg_mtadv}
    \KwIn{\textit{Target:} face image \(X_1,...,X_N\), deep learning model \(f_1(\cdot),...,f_M(\cdot)\); \textit{Source:} face image $X^{source}$; \textit{Settings:} perturbation size $\epsilon$, step size $\alpha$, maximum steps $t_{max}$, convergence threshold $\tau_{conv}$, system thresholds \(\tau_1,...,\tau_M\)}
    \KwOut{Adversarial example $X^{adv}$}
    \BlankLine
    \Begin(Initialization){\label{alg_Initialization_start}
        \(\delta^0\sim U(-\epsilon,\epsilon)\)\\
        \(X^0:=X^{source}+\delta^0\)\label{alg_Initialization_end}
        }
    \Repeat{Satisfy \eqref{eq_convergence_c1} or \eqref{eq_convergence_c2} or \(t=t_{max}\)}{\label{alg_loop_start}
        \(J_{MT}(X^{adv})=\Sigma_{m=1}^{M}(\frac{\Sigma_{n=1}^{N}||f_m(X^{adv})-f_m(X_n)||}{\tau_m})\)\ \eqref{eq_loss_mtadv} \label{alg_line_loss}\\
        \(X^{t+1}=Clip_{X^{source},\epsilon}\{X^{t}+\alpha \cdot sign(\nabla_{X^{t}}J_{ST})\}\)\label{alg_line_pgd}
        }\label{alg_loop_end}
    \Return{\(X^{adv}:=X^{t_{stop}}\)}\label{alg_line_return}
\end{algorithm}

The algorithm of MTADV is illustrated in Algorithm \ref{alg_mtadv}. It consists of a random initialization (Steps \ref{alg_Initialization_start}-\ref{alg_Initialization_end}) and an iterative optimization (Steps \ref{alg_loop_start}-\ref{alg_loop_end}). The optimization will be aborted when any of the following stop criteria are satisfied (Step \ref{alg_loop_end}):
\begin{itemize}
    \item achieving the maximum number of steps: \(t=t_{max}\). Note that a large \(t_{max}\) can exhaust the attack, yet a small \(t_{max}\) may leave the optimization unconverged.
    \item achieving convergence; i.e., the change in loss value is slim. The learning may achieve convergence prior to reaching \(t_{max}\). To evaluate the convergence or divergence, the variation in loss value $\Delta$ between continuous steps is first recorded as
    \begin{equation}
        \begin{aligned}
            &\Delta J^{t+1}=J(X^t,*)-J(X^{t+1},*),\\
            &S=\{\Delta J^{t},\Delta J^{t+1},\Delta J^{t+2},\Delta J^{t+3},\Delta J^{t+4}\},
        \end{aligned}
        \label{eq_convergence_delta}
    \end{equation}
    where $S$ is a set of $\Delta$ with the latest five steps. Then, this stop criterion is defined as 
    \begin{equation}
        \forall\Delta\in S,|\Delta|\leq\tau_{conv},
        \label{eq_convergence_c1}
    \end{equation}
    where $\tau_{conv}$ is a pre-defined convergence threshold;
    \item achieving settlement within an error range around a value; i.e., the loss value increases and decreases regularly. Thus, the stop criterion is defined as
    \begin{equation}
        \exists\Delta^*,\Delta^{**}\in S, \Delta^*\leq 0 \wedge \Delta^{**}\leq 0.
        \label{eq_convergence_c2}
    \end{equation}
\end{itemize}
Ultimately, the perturbed example satisfying the stop criteria is outputted as the adversarial example $X^{adv}$ (Step \ref{alg_line_return}).
\section{Experiments}
\label{Experiments}
\subsection{Experimental settings}
\label{settings}
\subsubsection{Datasets}
To evaluate the generalization of MTADV across various data, e.g., attacking a target user in Dataset A using a source image in Dataset B, three face datasets, i.e., LFW \cite{Hua08}, CelebA \cite{Liu15}, and CelebA-HQ \cite{karras2018progressive}, are adopted for experiments. Apart from the ablation study in image size, the target images are from the CelebA dataset, and the source images are from the LFW dataset. Particularly for the ablation study in image size, a high-quality version ($1024\times1024$) of the CelebA dataset, namely CelebA-HQ \cite{karras2018progressive}, is adopted for source images. In contrast, the target images are from the LFW dataset. The source images are then resized to various qualities from $1000\times1000$ to $112\times112$. From all three datasets, the subjects with more than ten images are chosen, which results in 158 subjects in the LFW dataset and 922 subjects in the CelebA dataset. Then, the first ten face images of each subject are selected. Eventually, to pair the source and target images, 1,580 (\(10\times158\)) samples are selected from each dataset.

\subsubsection{Deep learning models for face authentication}
We conduct experiments using three deep learning models for face authentication, i.e., FaceNet \cite{Sch15}, InsightFace \cite{Den19}, and CurricularFace \cite{huang2020curricularface}. Face images are converted into 512-dimensional facial features after feature extraction.

\subsubsection{Benchmark single-task adversarial attacks}
\label{CBCE}
We conducted experiments to evaluate the feasibility of MTADV against the latest deep learning model for face authentication, i.e., CurricularFace \cite{huang2020curricularface}, in the classic single-task attack scenarios and compare the results with existing benchmark single-task adversarial attacks as listed in Table \ref{tab_attack}. We have also conducted a comparison between MTADV and the most recent state-of-the-art black-box single-task attack, the sibling-attack \cite{li2023sibling}. Please note that since they have not released the source code, this comparison was performed using the same dataset (LFW) and deep learning model (FaceNet - IR architecture), with the numerical results directly cited from their paper. Note that it is not appropriate to compare MTADV with existing methods in the multi-task scenarios as MTADV is the only method applicable to multi-task attacks.
\begin{table}[!htb]
	\setlength{\abovecaptionskip}{0.1cm}
	\setlength{\belowcaptionskip}{0.1cm}
    \caption{Settings of the adversarial attacks}
    \label{tab_attack}
    \centering
    \setlength{\tabcolsep}{4mm}{\begin{tabular}{cc}
        \hline
        Technique&Settings\\
        \hline
        \hline
        FGSM \cite{Goo14}&$\epsilon=0.03$\\
        DeepFool \cite{Moo16}&$\epsilon=0.03$, $t_{max}=40$, overshoot $\eta=0.02$\\
        CW \cite{Car171}&$\epsilon=0.03$, $\alpha=0.001$, $t_{max}=1000$, binary search iterations = 20\\
        PGD \cite{madry2018towards}&$\epsilon=0.03$, $\alpha=0.001$, $t_{max}=40$\\
        SGADV \cite{wang2021fg}&$\epsilon=0.03$, $\alpha=0.001$, $t_{max}=1000$, $\tau_{conv}=0.0001$\\
        \hline
        MTADV (ours)&$\epsilon=0.03$, $\alpha=0.001$, $t_{max}=1000$, $\tau_{conv}=0.0001$\\
        \hline
    \end{tabular}}
\end{table}

\subsubsection{Evaluation metrics}
\begin{itemize}
	\item \textit{Equal error rate (EER)} and \textit{threshold} are computed when the system is set up. The attack success rate is expected to be significantly greater than EER, and the dissimilarity should not exceed the threshold.
	\item \textit{False positive rate (FPR)} and \textit{true positive rate (TPR)} are computed to obtain the receiver operating characteristic (ROC) curve showing the security under the adversarial attack.
	\item \textit{Attack success rate (ASR)} refers to the ratio of adversarial examples that successfully access the system. For an adversarial example $X^{adv}$ impersonating a target user $X$ with $N$ face images, ASR is computed as:
	\begin{equation}
	    ASR = \frac{1}{N}\Sigma_{i=1}^{N}(||f(X^{adv}-f(X^i)||\leq\tau).
	\end{equation}
	ASR is the most critical indicator of security and attack performance. A higher ASR indicates better attack performance and poorer security.
	\item \textit{Dissimilarity} indicates the closeness between the adversarial example and the target image in the feature space. An attack is successful when the dissimilarity is not higher than the threshold. In this paper, \(dissimilarity\in[0,1]\) is computed by the normalized cosine distance:
	\begin{equation}
	    dissimilarity=0.5\times cos(f(X^{adv}),f(X))+0.5.
	\end{equation}
	\item The \textit{Structural Similarity Index Measure (SSIM)} \cite{wang2004image} is a quantitative measurement to evaluate the impact of the perturbation (i.e., the image quality of the adversarial example). An average high value is expected whenever the adversarial example is very similar to its source image, thus allowing to evade human inspection.
	\item The \textit{Learned Perceptual Image Patch Similarity (LPIPS)} \cite{zhang2018perceptual} is another image-based metric to measure the difference in pixel intensities between corresponding image patches. A high value is diagnostic of a significant difference in the intensity values of image patches from the adversarial example and its source image.
\end{itemize}

\subsubsection{Implementation}
We adopt a publicly available tool, namely Foolbox \cite{rauber2020foolbox,rauber2017foolbox}, for the experimental implementation, using PyTorch. The machine used for the simulations is equipped with i7-9700 CPU, 64GB RAM, and NVIDIA TITAN Xp GPU.

\subsection{Performance of MTADV}
\label{performance}

\subsubsection{Attack success rates in five attack scenarios}
\label{performance_asr}
In this section, the success of MTADV in the investigated five attacks scenarios, i.e., ST, MA, UA, TA, and CA, is  experimentally demonstrated in terms of ASRs.

\textbf{Single-task attack (MTADV-ST)}:
Table \ref{tab_ST} presents the feasibility of MTADV-ST to three different deep learning models. Specifically, MTADV-ST achieves 100\% ASRs in the white-box setting and 93.23\%-98.74\% ASRs in the gray-box setting. Compared between the columns of ``Gray-box'' and ``EER'', it is observed that the gray-box attack ASRs are highly correlated with the clean system performance for authentication (\(ASR+EER\approx 100\%\)). This is reasonable because the adversarial example generated by MTADV-ST is extremely close to the target image in the feature space in terms of the dissimilarity score (referring to the column of ``Dissimilarity''), so they can be regarded as the same image sharing the system error rate. Hence, we hold the belief that further improvement in performance may not be theoretically achievable due to the constraints of the system. In addition, the comparison results in Table \ref{tab_ST_compare} show that MTADV-ST dramatically outperforms all white-box benchmark adversarial attacks (ASR raises from 22.59\% to 94.88\%) and gains competitive performance with the best gray-box attack. However, these benchmark adversarial attacks are not applicable to multi-task attack scenarios. Furthermore, the results presented in Table \ref{tab_ST_sibling} indicate that within the context of our threat models, we can achieve even better performance than the state-of-the-art black-box attacks, such as the sibling-attack \cite{li2023sibling}.
\begin{table}[!b]
    \setlength{\abovecaptionskip}{0.1cm}
	\setlength{\belowcaptionskip}{0.1cm}
    \centering
    \begin{threeparttable}
        \caption{Single-task attack performance (MTADV-ST)}
        \label{tab_ST}
        \setlength{\tabcolsep}{0.2mm}{\begin{tabular}{ccccccccc}
            \hline
            \multirow{2}{*}{Deep learning model}&\multirow{2}{*}{$\tau$}&\multirow{2}{*}{EER (\%)}&\multicolumn{2}{c}{ASR (\%)}&\multirow{2}{*}{Dissimilarity}&\multirow{2}{*}{SSIM}&\multirow{2}{*}{LPIPS}&Time cost (s)\\
            &&&White box&Gray box&&&&per example\\
            \hline
            FaceNet&0.2967&1.20&100.00&98.74&0.0032&0.8551&0.0984&2.86\\
            InsightFace&0.4146&6.23&100.00&93.23&0.0665&0.8746&0.0458&4.23\\
            CurricularFace&0.4278&4.41&100.00&94.88&0.0794&0.8790&0.0410&4.89\\
            \hline
        \end{tabular}}
    \end{threeparttable}
\end{table}
\begin{table}[!htb]
	\setlength{\abovecaptionskip}{0.1cm}
	\setlength{\belowcaptionskip}{0.1cm}
    \centering
        \caption{Comparison among MTADV-ST and other adversarial attacks against CurricularFace in single-task attack scenarios}
        \label{tab_ST_compare}
        \setlength{\tabcolsep}{0.3mm}{\begin{tabular}{ccccccccc}
            \hline
            \multirow{2}{*}{Attack}&\multirow{2}{*}{$\tau$}&\multirow{2}{*}{EER (\%)}&\multicolumn{2}{c}{ASR (\%)}&\multirow{2}{*}{Dissimilarity}&\multirow{2}{*}{SSIM}&\multirow{2}{*}{LPIPS}&Time cost (s)\\
            &&&White box&Gray box&&&&per example\\
            \hline
            \hline
            FGSM&\multirow{5.5}{*}{0.4278}&\multirow{5.5}{*}{4.41}&59.62&22.59&0.4160&0.8213&0.0841&0.65\\
            DeepFool&&&100.00&11.32&0.4259&0.9971&0.0004&0.67\\
            CW&&&99.81&10.75&0.4277&0.9998&0.0001&112.56\\
            PGD&&&100.00&16.50&0.4184&0.9161&0.0127&0.55\\
            SGADV&&&100.00&94.86&0.0802&0.8785&0.0412&4.63\\
            \hline
            MTADV-ST (ours)&0.4278&4.41&100.00&94.88&0.0794&0.8790&0.0410&4.89\\
            \hline
        \end{tabular}}
\end{table}
\begin{table}[!htb]
	\setlength{\abovecaptionskip}{0.1cm}
	\setlength{\belowcaptionskip}{0.1cm}
    \centering
        \caption{Comparison between MTADV-ST and sibling-attack \cite{li2023sibling} against FaceNet in the white-box setting}
        \label{tab_ST_sibling}
        \setlength{\tabcolsep}{15mm}{\begin{tabular}{ccc}
            \hline
            Attack&ASR (\%)&SSIM\\
            \hline
            \hline
            Sibling-attack&97.6&0.626\\
            MTADV-ST (ours)&100&0.855\\
            \hline
        \end{tabular}}
\end{table}

\textbf{Morphing attack (MTADV-MA)}:
The results in Table \ref{tab_MA} demonstrate MTADV indeed applicable to multi-task attack scenarios. In this fundamental multi-task scenario impersonating multiple users, MTADV gains 100\% ASRs in the white-box setting and 88.51\%-91.83\% ASRs in the gray-box setting. These results reflect the geometric proof in Figure \ref{fig_geometry}(b). However, when comparing the gray-box results in Table \ref{tab_MA} with \ref{tab_ST}, the ASRs of MTADV-MA are slightly lower than those of MTADV-ST, which reflects the theoretically analysis in Section \ref{proof} about the deterioration due to the different closeness.
\begin{table}[!htb]
    \setlength{\abovecaptionskip}{0.1cm}
	\setlength{\belowcaptionskip}{0.1cm}
    \centering
    \begin{threeparttable}
        \caption{Morphing attack performance (MTADV-MA)}
        \label{tab_MA}
        \setlength{\tabcolsep}{0.2mm}{\begin{tabular}{ccccccccc}
            \hline
            \multirow{2}{*}{Deep learning model}&\multirow{2}{*}{$\tau$}&\multirow{2}{*}{EER (\%)}&\multicolumn{2}{c}{ASR (\%)}&\multirow{2}{*}{Dissimilarity}&\multirow{2}{*}{SSIM}&\multirow{2}{*}{LPIPS}&Time cost (s)\\
            &&&White box&Gray box&&&&per example\\
            \hline
            FaceNet&0.2967&1.20&100.00&88.51&0.1445&0.8550&0.0978&3.14\\
            InsightFace&0.4146&6.23&100.00&89.96&0.1919&0.8755&0.0447&4.22\\
            CurricularFace&0.4278&4.41&100.00&91.83&0.2064&0.8799&0.0399&3.34\\
            \hline
        \end{tabular}}
    \end{threeparttable}
\end{table}

\textbf{Universal attack (MTADV-UA)}:
The universal adversarial attack aims to generate an ``average face'' by learning an adversarial example based on 50 subjects. We further evaluate two sub-scenarios by requiring 1 and 10 face images from each subject, respectively. Then, this universal adversarial example is fed into a target authentication system to attack other 108 subjects (we have 158 subjects in the database). In this universal attack, MTADV does not require any target image. The results presented in Table \ref{tab_UA} indicate that MTADV-UA is effective to act as a universal attack. For example, in the system using Insightface, the universal adversarial example can match 39.44\% users, although this adversarial example is learnt referring to none of them. Moreover, MTADV-UA can measure the robustness of different deep learning models. From the results in Table \ref{tab_UA}, MTADV uncovers that InsightFace and CurricularFace suffer poor security and robustness compared with FaceNet. This observation aligns with the analysis presented in Section \ref{proof}, which indicates that a higher threshold is more susceptible to universal attacks. In addition, the results indicate that learning the universal adversarial example with more face images from each subject leads to better performance. In practice, this condition is easy to achieve by using a public dataset.
\begin{table}[!tb]
    \setlength{\abovecaptionskip}{0.1cm}
	\setlength{\belowcaptionskip}{0.1cm}
    \centering
    \begin{threeparttable}
        \caption{Universal attack performance (MTADV-UA)}
        \label{tab_UA}
        \setlength{\tabcolsep}{0.2mm}{\begin{tabular}{cccccccc}
            \hline
            \multirow{2}{*}{Deep learning model}&\multicolumn{3}{c}{Dataset$^{*}$}&\multicolumn{2}{c}{1 image for each subject}&\multicolumn{2}{c}{10 images for each subject}\\
            &Source&Learning&Attack&ASR (\%)&Time cost (s)&ASR (\%)&Time cost (s)\\
            \hline
            \hline
            \multirow{4}{*}{FaceNet}&\multirow{2}{*}{LFW}&CelebA&CelebA&5.09&5.71&5.37&4.82\\
            &&CelebA&LFW&0.65&5.79&0.74&5.42\\
            \Xcline{2-8}{0.4pt}
            &\multirow{2}{*}{CelebA}&LFW&LFW&6.20&9.30&7.31&8.44\\
            &&LFW&CelebA&0&12.78&0&8.78\\
            \hline
            \multirow{4}{*}{InsightFace}&\multirow{2}{*}{LFW}&CelebA&CelebA&19.17&5.58&20.19&4.31\\
            &&CelebA&LFW&15.83&4.71&18.70&4.21\\
            \Xcline{2-8}{0.4pt}
            &\multirow{2}{*}{CelebA}&LFW&LFW&31.76&5.61&39.44&6.42\\
            &&LFW&CelebA&22.41&6.22&31.48&5.64\\
            \hline
            \multirow{4}{*}{CurricularFace}&\multirow{2}{*}{LFW}&CelebA&CelebA&3.42&1.99&6.94&1.65\\
            &&CelebA&LFW&2.87&1.61&3.70&2.44\\
            \Xcline{2-8}{0.4pt}
            &\multirow{2}{*}{CelebA}&LFW&LFW&16.67&6.54&24.17&6.58\\
            &&LFW&CelebA&10.83&5.11&18.61&6.10\\
            \hline
        \end{tabular}}
        \begin{tablenotes}
          \footnotesize
          \item $^{*}$ The subjects in ``Source'' and ``Learning'' are not involved in ``Attack."
        \end{tablenotes}
    \end{threeparttable}
\end{table}

\textbf{Transferable attack (MTADV-TA)}:
TA is a basic multi-task attack scenario for attacking multiple systems. The results in Table \ref{tab_TA} demonstrate the effectiveness of MTADV-TA. The adversarial example, which can gain illegal access to System A, can also hack into System B with a minimum 91.7\% probability.
\begin{table}[!htb]
    \setlength{\abovecaptionskip}{0.1cm}
	\setlength{\belowcaptionskip}{0.1cm}
    \centering
    \begin{threeparttable}
        \caption{Transferable attack performance (MTADV-TA)}
        \label{tab_TA}
        \setlength{\tabcolsep}{0.2mm}{\begin{tabular}{ccccccccc}
            \hline
            \multirow{2}{*}{Deep learning model$^*$}&\multirow{2}{*}{$\tau$}&\multirow{2}{*}{EER (\%)}&\multicolumn{2}{c}{ASR (\%)}&\multirow{2}{*}{Dissimilarity}&\multirow{2}{*}{SSIM}&\multirow{2}{*}{LPIPS}&Time cost (s)\\
            &&&White box&Gray box&&&&per example\\
            \hline
            \hline
            FaceNet&0.2967&1.20&99.75&92.26&0.0851&\multirow{2}{*}{0.8659}&\multirow{2}{*}{0.0473}&\multirow{2}{*}{10.05}\\
            InsightFace&0.4146&6.23&100.00&91.70&0.1283&&&\\
            \hline
            InsightFace&0.4146&6.23&100.00&91.74&0.1305&\multirow{2}{*}{0.8731}&\multirow{2}{*}{0.0429}&\multirow{2}{*}{7.48}\\
            CurricularFace&0.4278&4.41&100.00&94.02&0.1311&&&\\
            \hline
            FaceNet&0.2967&1.20&99.81&92.45&0.0832&\multirow{2}{*}{0.8671}&\multirow{2}{*}{0.0447}&\multirow{2}{*}{6.57}\\
            CurricularFace&0.4278&4.41&100.00&93.90&0.1296&&&\\
            \hline
        \end{tabular}}
        \begin{tablenotes}
          \footnotesize
          \item $^{*}$ MTADV-TA adopts a single adversarial example to attack two deep learning models simultaneously.
        \end{tablenotes}
    \end{threeparttable}
\end{table}

\textbf{Counterattack (MTADV-CA)}:
A counterattack refers to an attack that attempts to defeat the system defended. In our context, MTADV-CA regards the defense module as another system apart from the feature extractor. The results in Table \ref{tab_CA} show the effectiveness of MTADV-CA against two latest adversarial defenses: an auto-encoder-based defense, perturbation inactivation (PIN) \cite{ren2022perturbation} and the state-of-the-art defense DiffPure \cite{nie2022DiffPure}. The results show that the single-task similarity-based gray-box adversarial attack (SGADV) successfully breaks the clean system without any defense (100\% ASR). However, these 100\% feasible adversarial examples are easily defended by PIN and DiffPure (ASRs drop to 19.71\% and 5\%, respectively). Nevertheless, MTADV-CA defeats PIN and DiffPure by raising ASRs to 94.97\% and 98.4\%, respectively.
\begin{table}[!htb]
    \setlength{\abovecaptionskip}{0.1cm}
	\setlength{\belowcaptionskip}{0.1cm}
    \centering
    \begin{threeparttable}
        \caption{Counterattack performance (MTADV-CA)}
        \label{tab_CA}
        \setlength{\tabcolsep}{4.5mm}{\begin{tabular}{cccccc}
            \hline
            Attack&EER (\%)&Defense&Counterattack?&ASR (\%)\\
            \hline
            SGADV&0.35&No&Not applicable&100.00\\
            SGADV&10.31&PIN \cite{ren2022perturbation}&No&19.71\\
            SGADV&5.00&DiffPure \cite{nie2022DiffPure}&No&5.00\\
            MTADV-CA (ours)&10.31&PIN \cite{ren2022perturbation}&Yes&94.97\\
            MTADV-CA (ours)&5.00&DiffPure \cite{nie2022DiffPure}&Yes&98.40\\
            \hline
        \end{tabular}}
    \end{threeparttable}
\end{table}

\subsubsection{Computational complexity}
\label{complexity}
In our experiments, the most time-consuming case is the universal attack against FaceNet in Table \ref{tab_UA}, which spends 12.78s GPU time per adversarial example generation. However, this efficiency is still practical for an attack. The major time-consuming steps in Algorithm \ref{alg_mtadv} are: 1) feature extraction \(f(\cdot)\), 2) loss computation (Step \ref{alg_line_loss}), and 3) gradient computation (Step \ref{alg_line_pgd}).

\subsubsection{Qualitative analysis}
The results of SSIM (higher than 0.855) and LPIPS (lower than 0.1) in Tables \ref{tab_ST}-\ref{tab_TA} show the satisfactory image quality of the adversarial examples in the setting \(\epsilon=0.03\). As observed in Fig. \ref{fig_ablation_epsilons_sample}, such results indicate that the differences between adversarial examples and source images are visually negligible so that they will evade human inspection.
\def\sizePerturbation{0.5in}
\begin{figure}[!htb]
 	\setlength{\abovecaptionskip}{0.1cm}
 	\setlength{\belowcaptionskip}{0.1cm}
    \centering
    \setlength{\tabcolsep}{0.1mm}
    \begin{tabular}{c|ccccccccc|c}
        &$\epsilon$&0.001&0.005&0.01&0.03&0.05&0.1&0.5&1&\\
        &&&&&&&&&\\
        \multirow{2}{*}{\begin{tabular}{c}\includegraphics[width=\sizePerturbation]{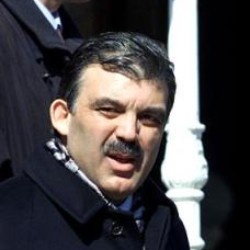}\\Source\end{tabular}}&\multirow{-3.5}{*}{Adv}&\includegraphics[width=\sizePerturbation]{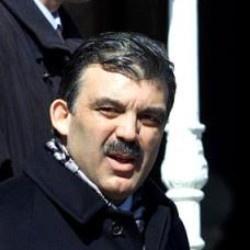}&\includegraphics[width=\sizePerturbation]{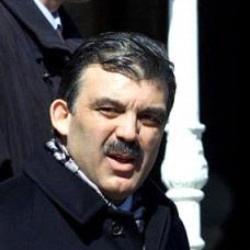}&\includegraphics[width=\sizePerturbation]{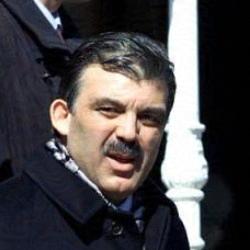}&\includegraphics[width=\sizePerturbation]{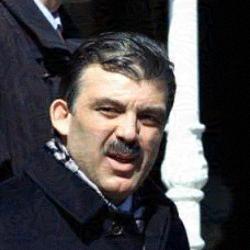}&\includegraphics[width=\sizePerturbation]{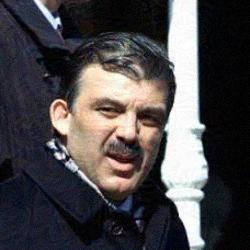}&\includegraphics[width=\sizePerturbation]{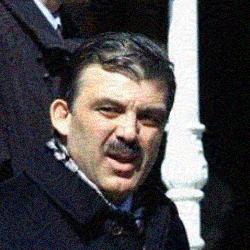}&\includegraphics[width=\sizePerturbation]{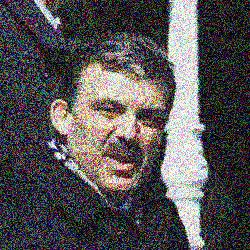}&\includegraphics[width=\sizePerturbation]{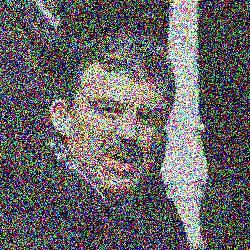}&\multirow{2}{*}{\begin{tabular}{c}\includegraphics[width=\sizePerturbation]{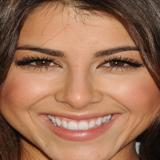}\\Target
        \end{tabular}}\\
        &\multirow{-3.5}{*}{Diff}&\includegraphics[width=\sizePerturbation]{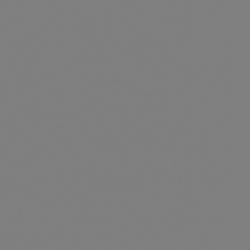}&\includegraphics[width=\sizePerturbation]{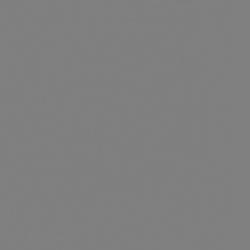}&\includegraphics[width=\sizePerturbation]{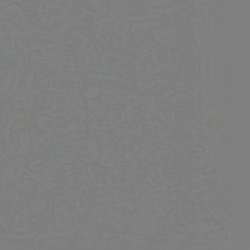}&\includegraphics[width=\sizePerturbation]{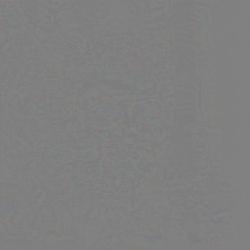}&\includegraphics[width=\sizePerturbation]{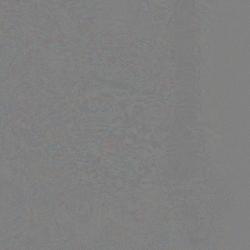}&\includegraphics[width=\sizePerturbation]{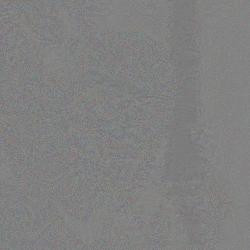}&\includegraphics[width=\sizePerturbation]{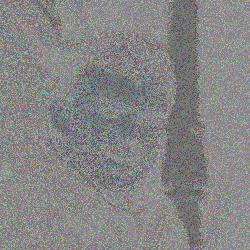}&\includegraphics[width=\sizePerturbation]{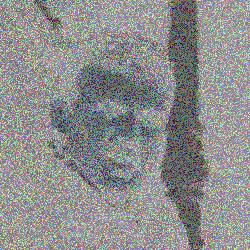}&\\
        &&&&&&&&&\\
        &SSIM&0.9996&0.9905&0.9676&0.8597&0.7306&0.4858&0.0911&0.0354&\\
        &LPIPS&0.0001&0.0053&0.0237&0.0902&0.1753&0.3812&1.2378&1.5055
        
    \end{tabular}
    \Description{Illustration of adversarial examples and differences in various perturbation sizes}
    \caption{Illustration of adversarial examples (Adv) and differences (Diff) in various perturbation sizes. As observed, the difference among the adversarial example is visually negligible when \(\epsilon\leq0.03\), while the perturbations are clearly visible when \(\epsilon>0.05\). Our setting is \(\epsilon=0.03\).}
    \label{fig_ablation_epsilons_sample}
\end{figure}

\subsection{Ablation study}
\label{ablation}
We conduct thorough ablation studies to showcase the impact of five parameters (settings) on the attack success rate, efficiency, and image quality: (a) the number of learning steps, (b) perturbation size, (c) step size, and (d) image size, (e) threshold.

\subsubsection{Number of learning steps}
\begin{figure}[!htb]
	\setlength{\abovecaptionskip}{0.1cm}
	\setlength{\belowcaptionskip}{0.1cm}
    \centering
    \subfloat[FaceNet]{\includegraphics[width=5.4in]{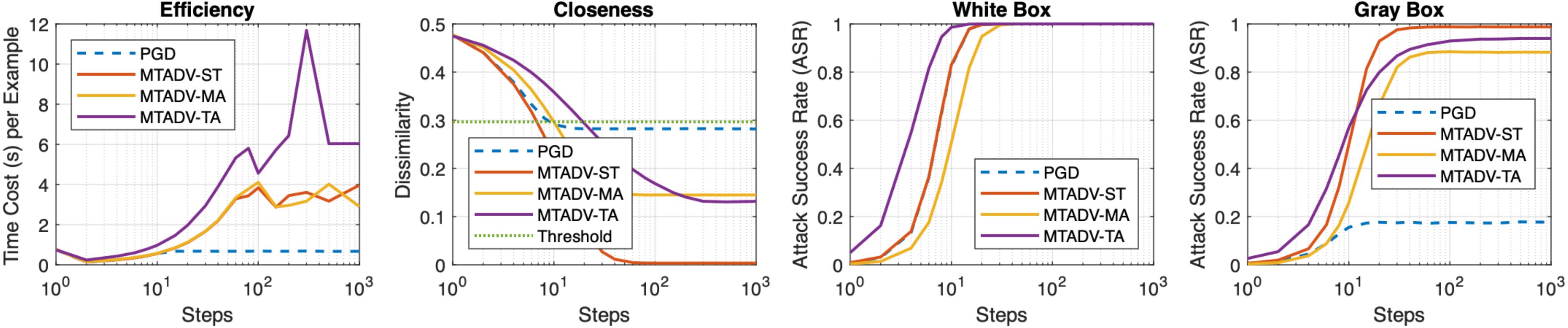}
    \label{fig_ablation_steps_fn}}
    
    \subfloat[InsightFace]{\includegraphics[width=5.4in]{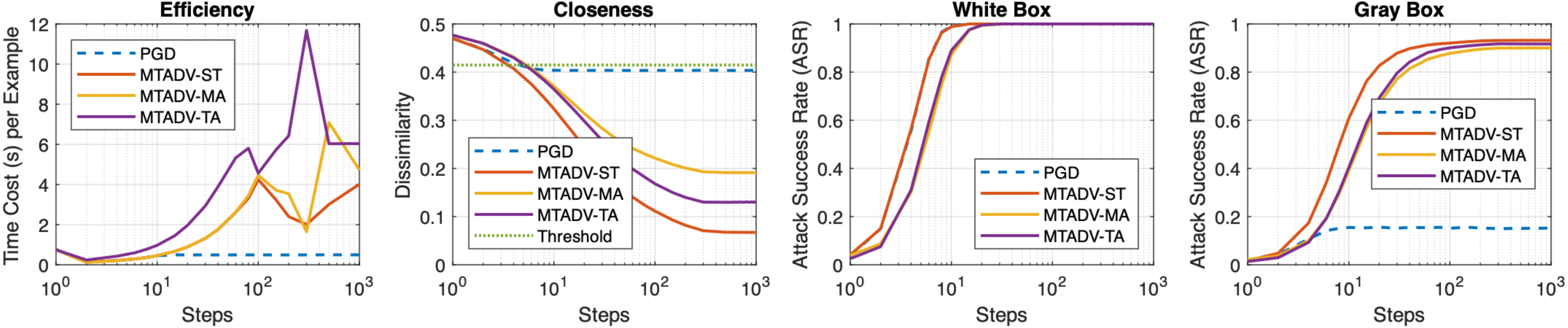}
    \label{fig_ablation_steps_if}} 
    
    \subfloat[CurricularFace]{\includegraphics[width=5.4in]{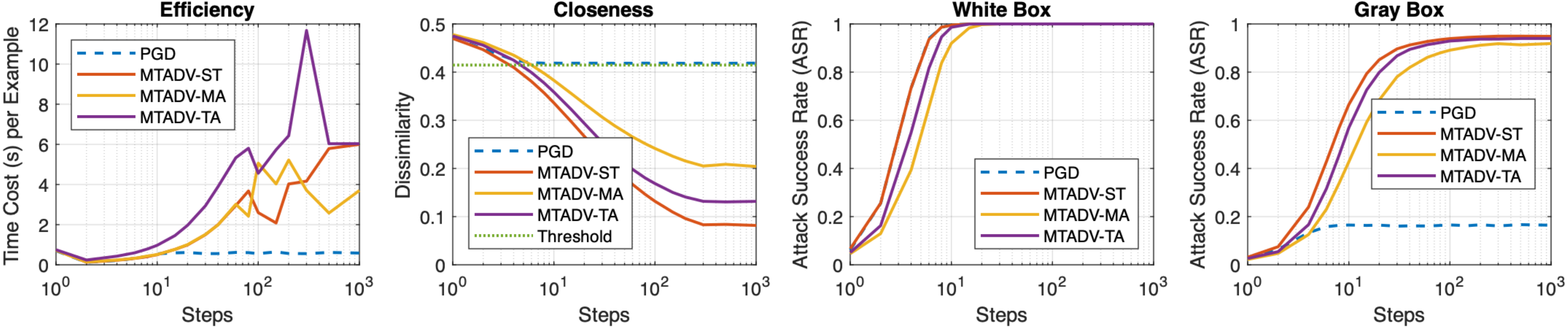}
    \label{fig_ablation_steps_cf}}
    \Description{The contribution of the number of learning steps to efficiency, attack performance, and closeness}
    \caption{The contribution of the number of learning steps to efficiency, attack performance, and closeness between the adversarial examples and the target images of the proposed MTADV, compared with PGD \cite{madry2018towards}, a representative of benchmark single-task attacks.}
    \label{fig_ablation_steps}
\end{figure}

Fig. \ref{fig_ablation_steps} illustrates the contribution of the number of learning steps to efficiency, attack performance, and closeness between the adversarial examples and target images. The results are also compared with PGD \cite{madry2018towards}, a representative of benchmark single-task attacks, as the number of learning steps results in a significant difference between MTADV and PGD. It is reasonable that the time cost increases and the dissimilarity decreases when the number of learning steps increases. However, due to the different algorithms, the dissimilarity under PGD attack converges when it just surpasses the threshold (green dotted line), while under MTADV attacks, the dissimilarity reduces until it cannot be significantly diminished. Thus, MTADV requires more learning steps to convergence than PGD, which is deemed less efficient. However, the additional time cost (number of learning steps) of MTADV leads to a significant ASR increase in the gray-box setting, as observed in the fourth column of Fig. \ref{fig_ablation_steps}. Note that the number of learning steps does not impact the ASR in the white-box setting after the dissimilarity decreases to below the threshold, as shown in the third column of Fig. \ref{fig_ablation_steps}.

\subsubsection{Perturbation size}
\label{ablation_perturbation}
Perturbation size contributes to efficiency, attack performance, and image quality. An overlarge perturbation size leads to computational expensiveness and results in apparent difference between the adversarial example and source image, while a tiny perturbation size produces poor ASR.

\begin{figure}[!htb]
	\setlength{\abovecaptionskip}{0.1cm}
	\setlength{\belowcaptionskip}{0.1cm}
    \centering
    \subfloat[FaceNet]{\includegraphics[width=5.4in]{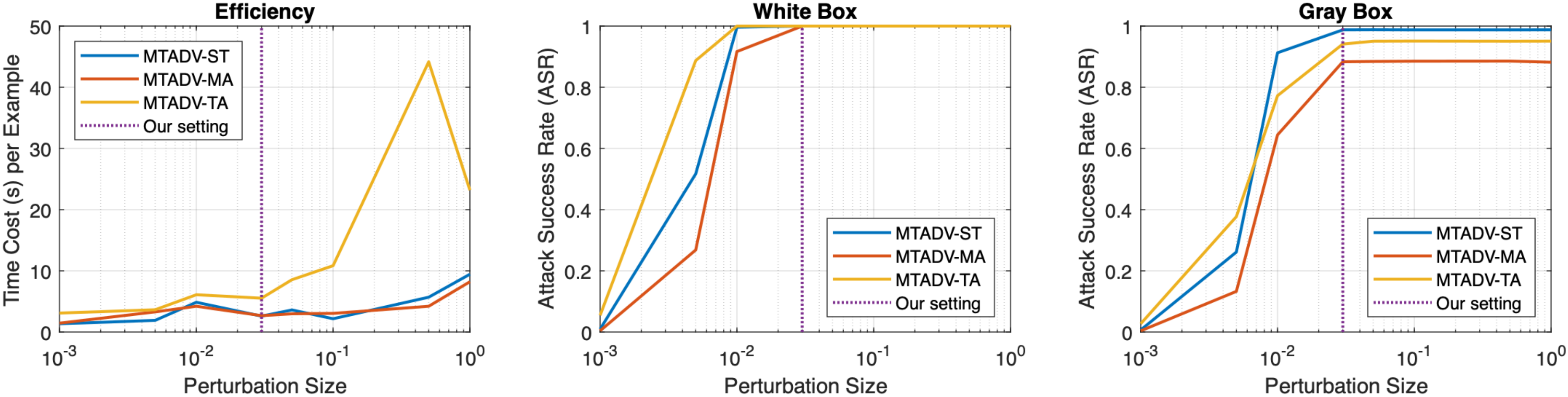}
    \label{fig_ablation_epsilons_fn}}
    
    \subfloat[InsightFace]{\includegraphics[width=5.4in]{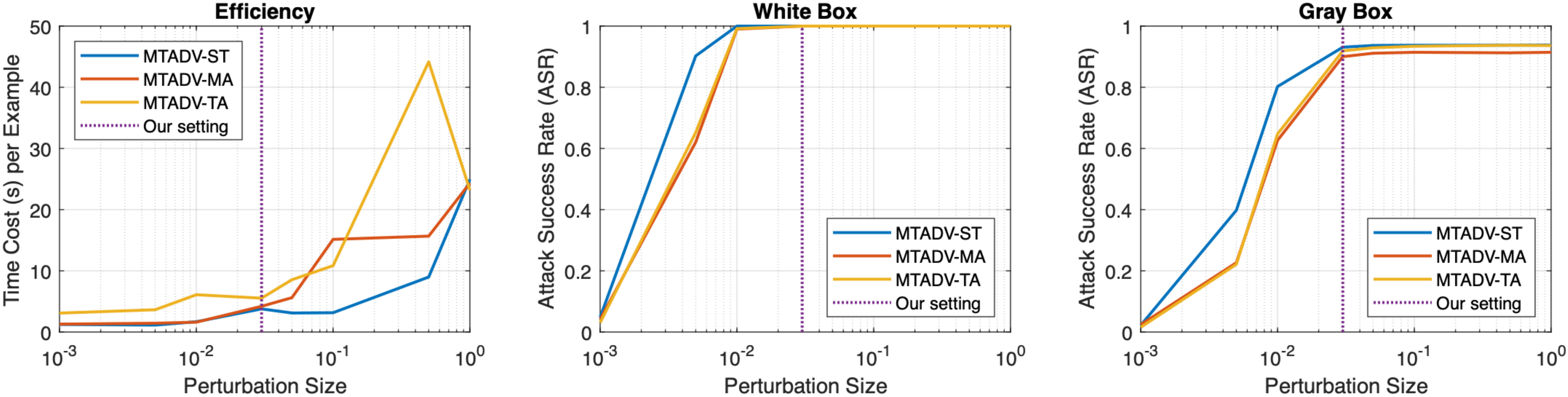}
    \label{fig_ablation_epsilons_if}}
    
    \subfloat[CurricularFace]{\includegraphics[width=5.4in]{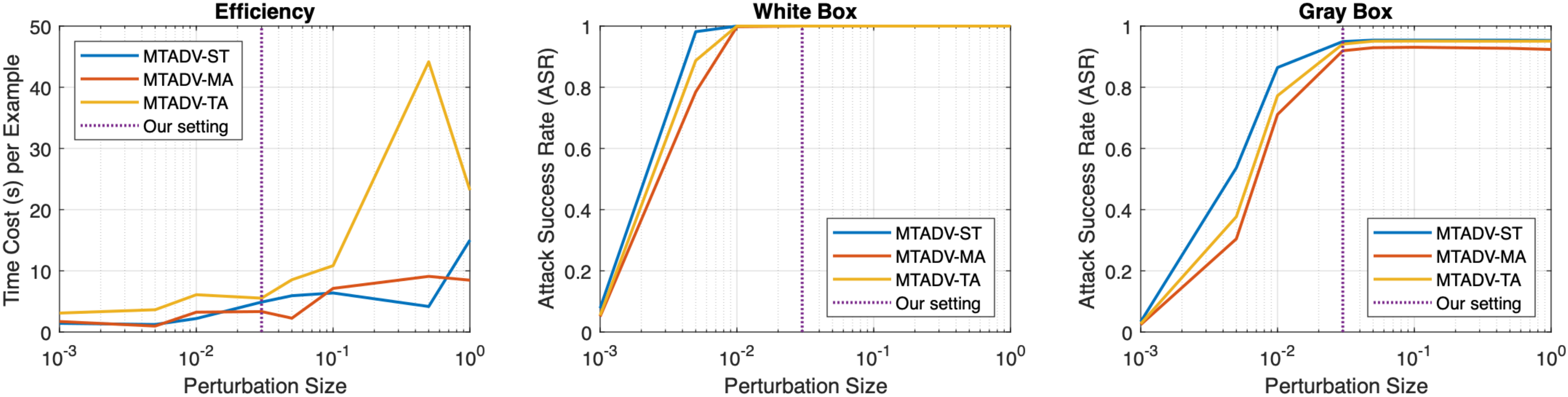}
    \label{fig_ablation_epsilons_cf}}
    \Description{The contribution of perturbation size to efficiency and attack performance of the proposed MTADV}
    \caption{The contribution of perturbation size to efficiency and attack performance of the proposed MTADV. The perturbation size should be at a medium level (e.g., $\epsilon=0.03$) to balance efficiency, attack performance, and image quality.}
    \label{fig_ablation_epsilons}
\end{figure}

Fig. \ref{fig_ablation_epsilons_sample} shows the adversarial examples and quantitative evaluation results (SSIM and LPIPS) in various perturbation sizes. The difference among the adversarial example is observed to be visually negligible when \(\epsilon\leq0.03\). At this stage, SSIM is greater than 0.85 and LPIPS is smaller than 0.01. However, as presented in Fig. \ref{fig_ablation_epsilons}, the ASR dramatically decreases when the perturbation size is too small as the change may not be significant enough to be extracted by the deep learning model. In contrast, the perturbations are clearly visible when \(\epsilon>0.05\), and the time cost dramatically rises when the perturbation size increases, as shown in the first column of Fig. \ref{fig_ablation_epsilons}. Therefore, the perturbation size should be at a medium level to balance efficiency, attack performance, and image quality.

\subsubsection{Step size}
\begin{figure}[!tb]
	\setlength{\abovecaptionskip}{0.1cm}
	\setlength{\belowcaptionskip}{0.1cm}
    \centering
    \subfloat[FaceNet]{\includegraphics[width=5.4in]{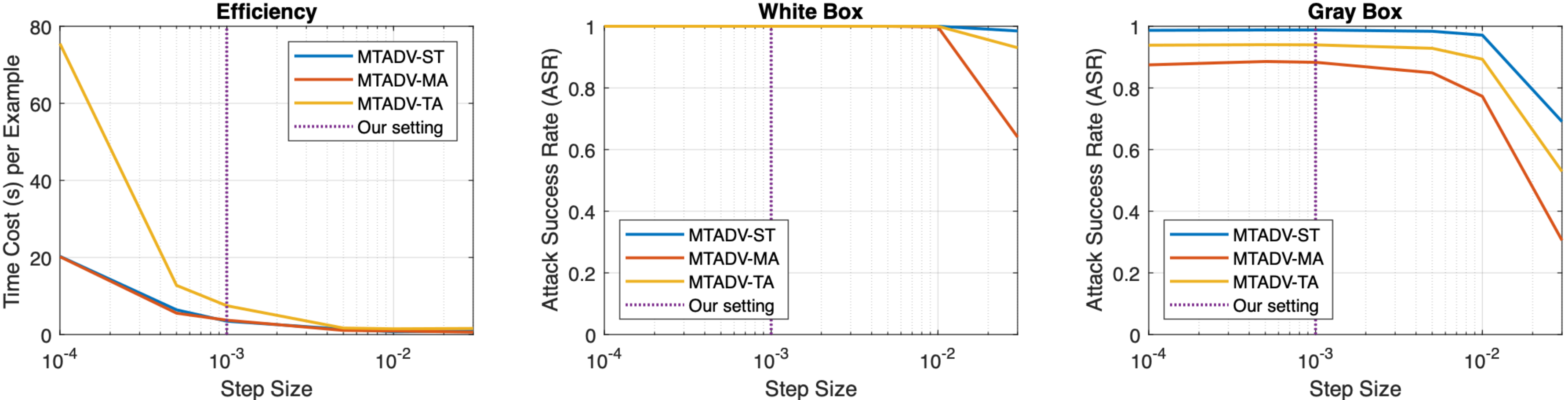}
    \label{fig_ablation_step_size_fn}}
    
    \subfloat[InsightFace]{\includegraphics[width=5.4in]{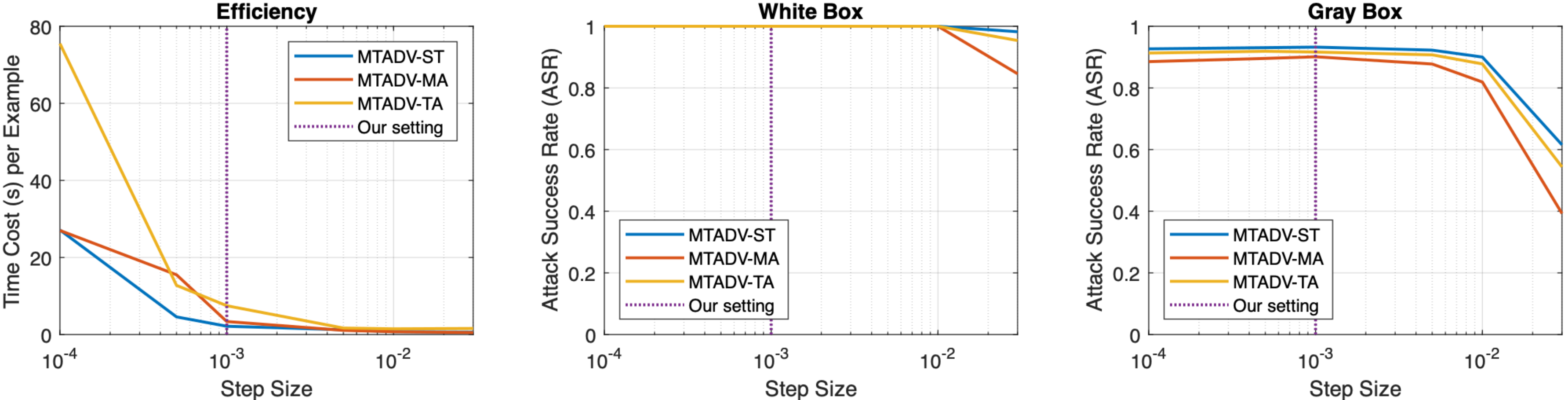}
    \label{fig_ablation_step_size_if}}
    
    \subfloat[CurricularFace]{\includegraphics[width=5.4in]{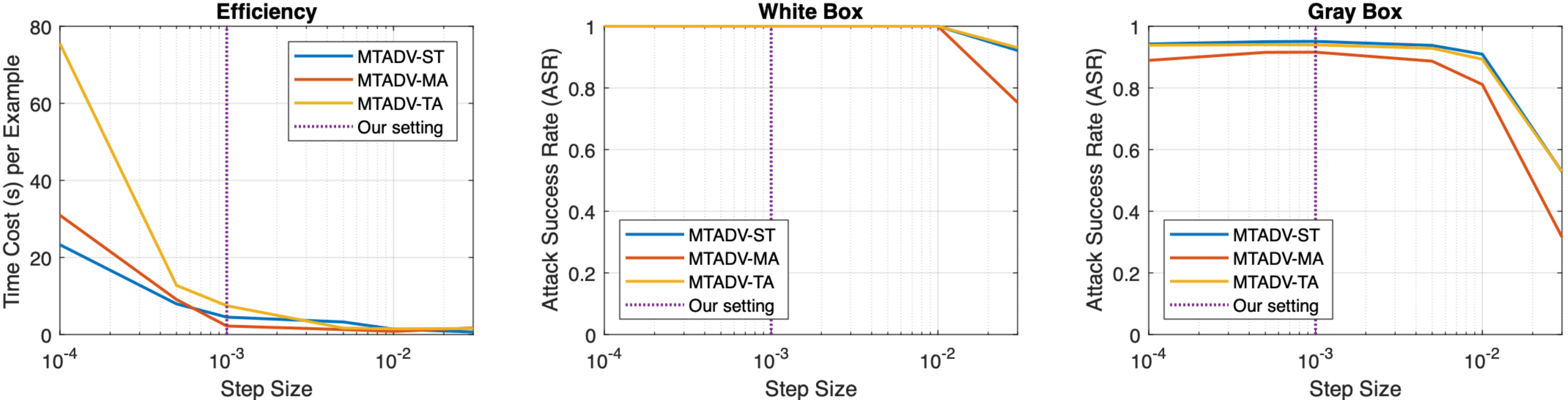}
    \label{fig_ablation_step_size_cf}}
    \Description{The contribution of step size to efficiency and attack performance of the proposed MTADV}
    \caption{The contribution of step size to efficiency and attack performance of the proposed MTADV. Like the perturbation size, the best setting for the step size is at a medium level, e.g., $\alpha=0.01$.}
    \label{fig_ablation_step_size}
\end{figure}

\begin{figure}[!tb]
	\setlength{\abovecaptionskip}{0.1cm}
	\setlength{\belowcaptionskip}{0.1cm}
    \centering
    \subfloat[FaceNet]{\includegraphics[width=4in]{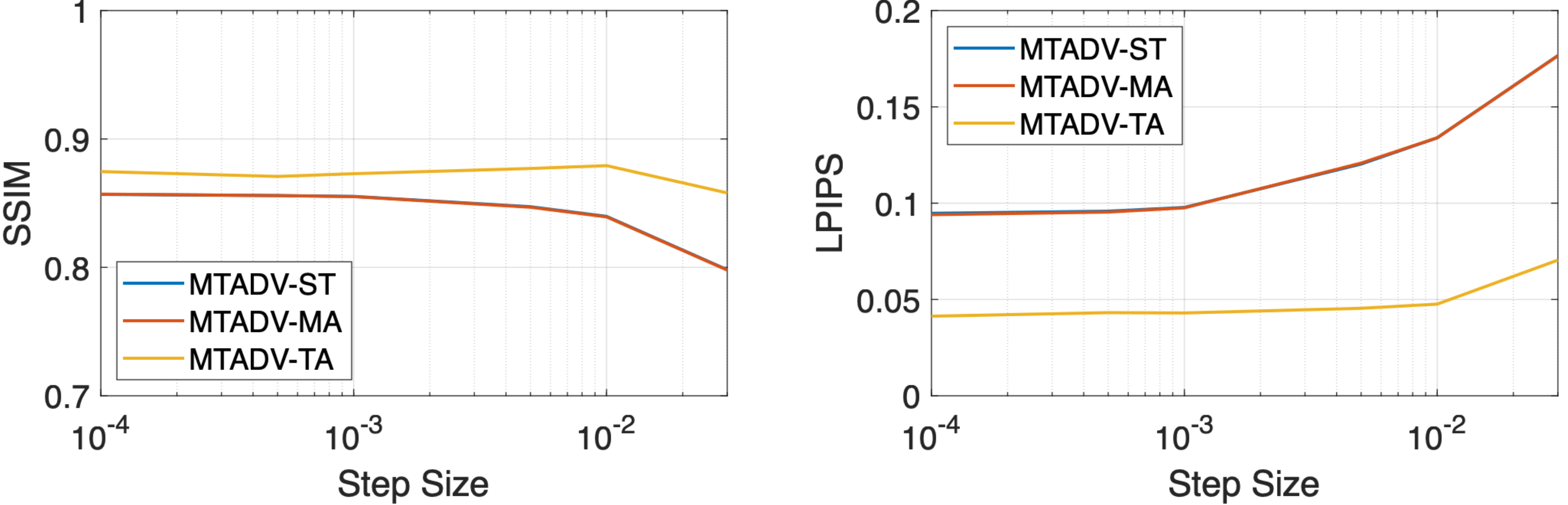}
    \label{fig_ablation_step_size_quality_fn}}
    
    \subfloat[InsightFace]{\includegraphics[width=4in]{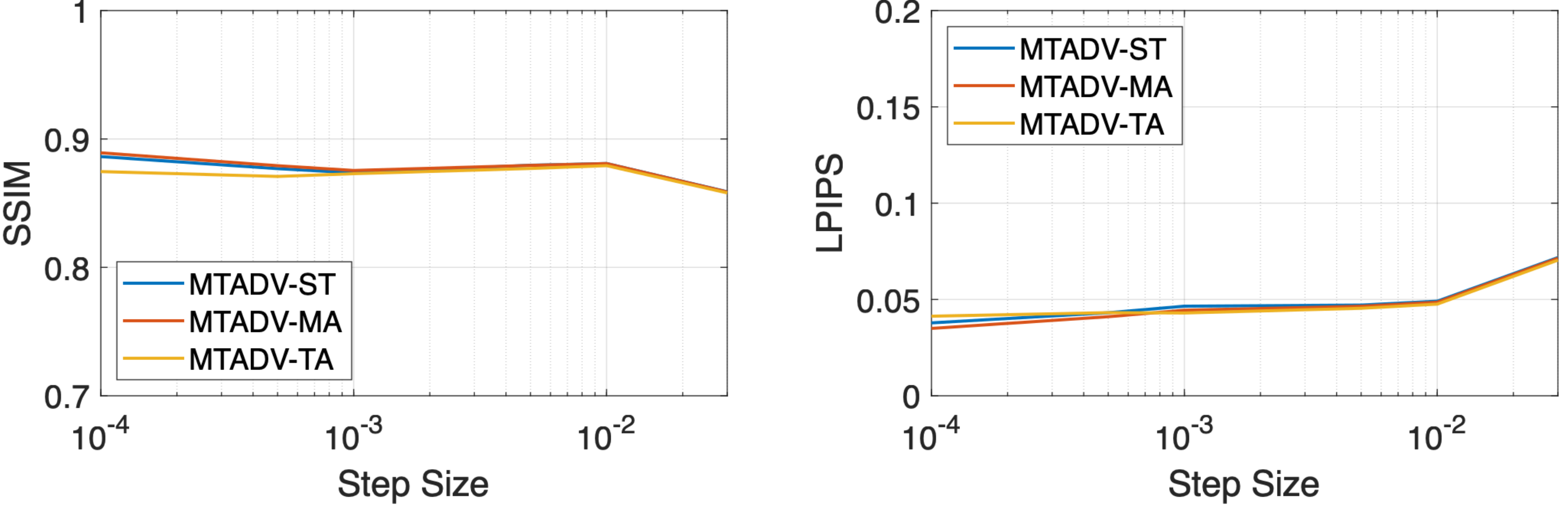}
    \label{fig_ablation_step_size_quality_if}}
    
    \subfloat[CurricularFace]{\includegraphics[width=4in]{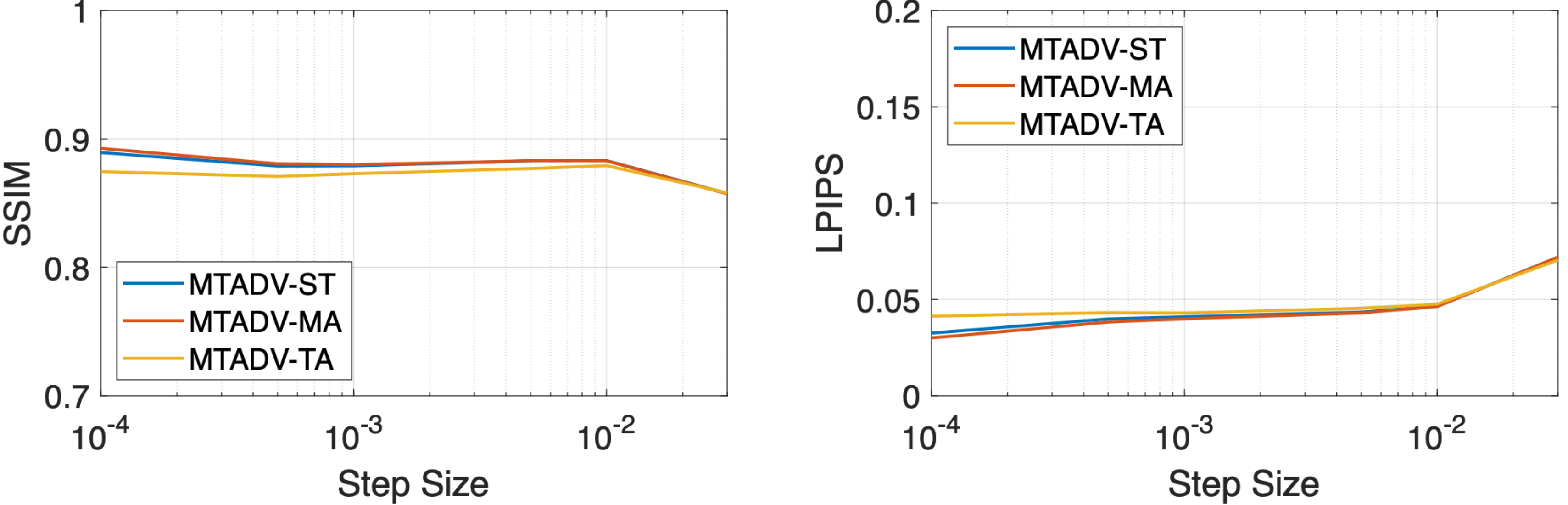}
    \label{fig_ablation_step_size_quality_cf}}
    \Description{Image quality in various step sizes}
    \caption{Image quality in various step sizes. The step size negligibly impacts the image quality of adversarial examples.}
    \label{fig_ablation_step_size_quality}
\end{figure}

Step size contributes to efficiency and attack performance yet image quality. As observed in Fig. \ref{fig_ablation_step_size}, the efficiency is significantly improved in a significant step size, while the ASRs under both white-box and gray-box attacks degrade if the step size is overlarge since it is difficult to achieve convergence if the change of each step is too significant. However, the image quality is not impacted by the step size, as shown in Fig. \ref{fig_ablation_step_size_quality}.

\subsubsection{Image size}
\label{image_size}
We investigate the contribution of input image size to efficiency and attack performance by varying the size of source images. CelebA-HQ dataset \cite{karras2018progressive} (with an original size of \(1024\times1024\)) is adopted as source images. Then, the source images are reshaped to \(1000\times1000\), \(500\times500\), \(250\times250\), \(160\times160\), and \(112\times112\), while the target images from LFW dataset are aligned to \(160\times160\) using multi-task cascaded convolutional neural networks \cite{zhang2016joint}. As shown in Table \ref{tab_ablation_imagesize}, the source images with a slightly larger size than the target images outperform other settings in terms of both efficiency and ASR. It further implies that a larger image contains more pixels that can be perturbed in the adversarial attack. Therefore, to balance attack performance and efficiency, the source images should be slightly larger than the target images. Note that this study is conducted by attacking only FaceNet instead of InsightFace and CurricularFace as a fixed input size (\(112\times112\) or \(224\times224\)) is required by these two models.
\begin{table}[!tb]
	\setlength{\abovecaptionskip}{0.1cm}
	\setlength{\belowcaptionskip}{0.1cm}
    \centering
        \caption{Attack performance using source images of various sizes}
        \label{tab_ablation_imagesize}
        \setlength{\tabcolsep}{2.5mm}{\begin{tabular}{ccccccc}
            \hline
            \multirow{3}{*}{Size}&\multicolumn{3}{c}{MTADV-ST}&\multicolumn{3}{c}{MTADV-MA}\\
            &\multirow{2}{*}{Time cost (s)}&\multicolumn{2}{c}{ASR (\%)}&\multirow{2}{*}{Time cost (s)}&\multicolumn{2}{c}{ASR (\%)}\\
            &&White box&Gray box&&White box&Gray box\\
            \hline
            $1000$&8.00&100.00&98.58&6.99&100.00&89.36\\
            $500$&2.85&100.00&98.62&3.04&100.00&89.79\\
            $250$&2.89&100.00&98.70&3.13&100.00&89.45\\
            $160$&5.48&100.00&98.21&5.68&99.97&83.21\\
            $112$&6.47&100.00&98.10&5.83&99.94&81.26\\
            \hline
        \end{tabular}}
\end{table}

\subsubsection{Threshold}
\begin{figure}[!tb]
	\setlength{\abovecaptionskip}{0.1cm}
	\setlength{\belowcaptionskip}{0.1cm}
    \centering
    \subfloat[FaceNet]{\includegraphics[width=4in]{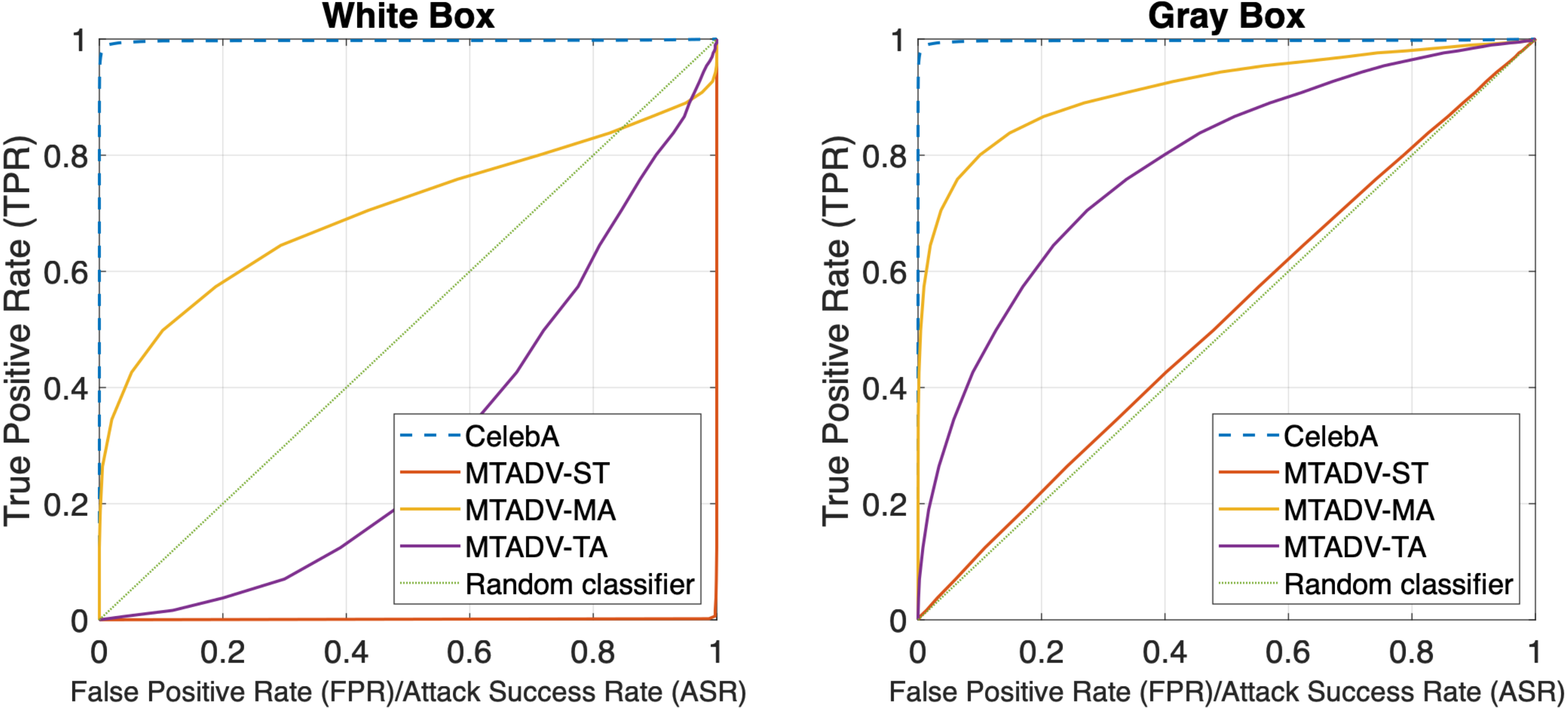}
    \label{fig_ablation_roc_fn}}
    
    \subfloat[InsightFace]{\includegraphics[width=4in]{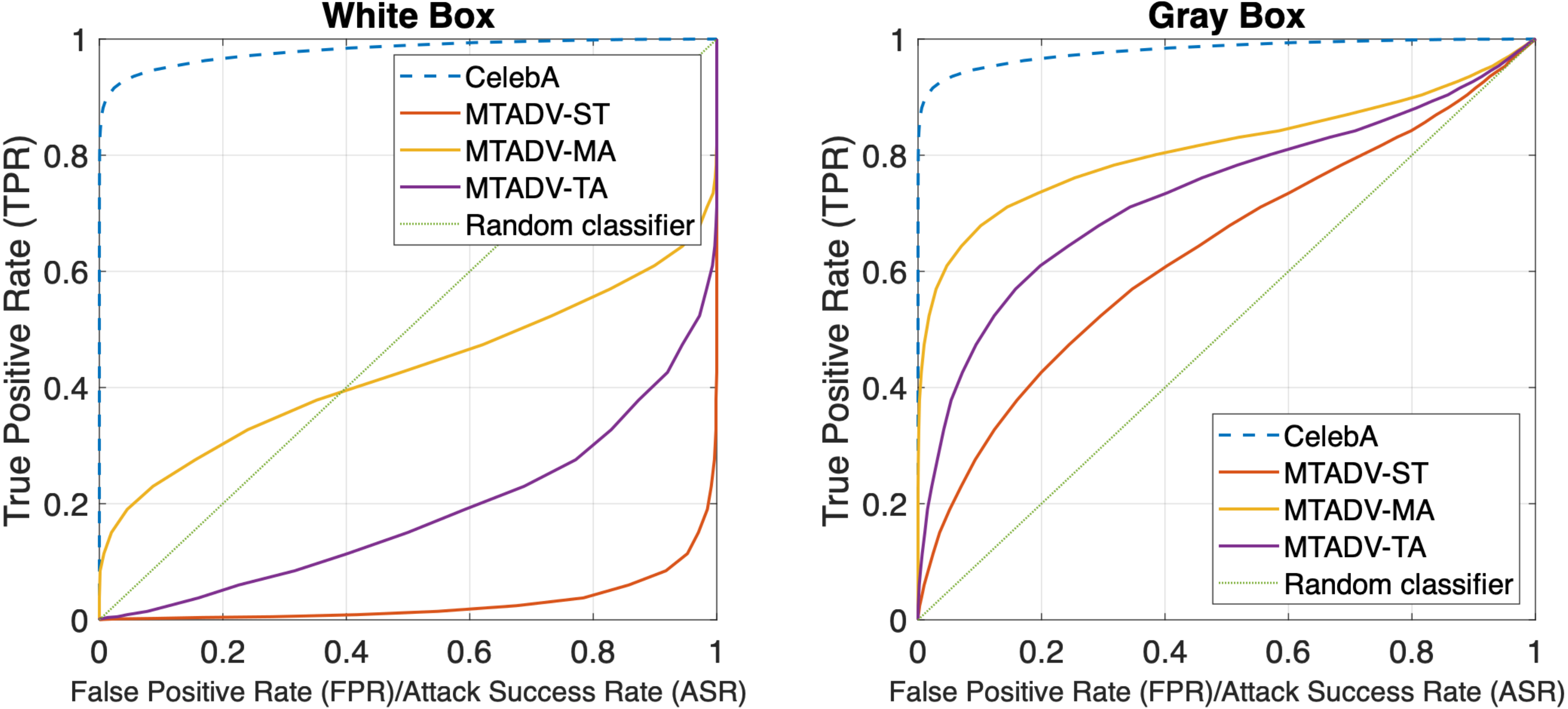}
    \label{fig_ablation_roc_if}}
    
    \subfloat[CurricularFace]{\includegraphics[width=4in]{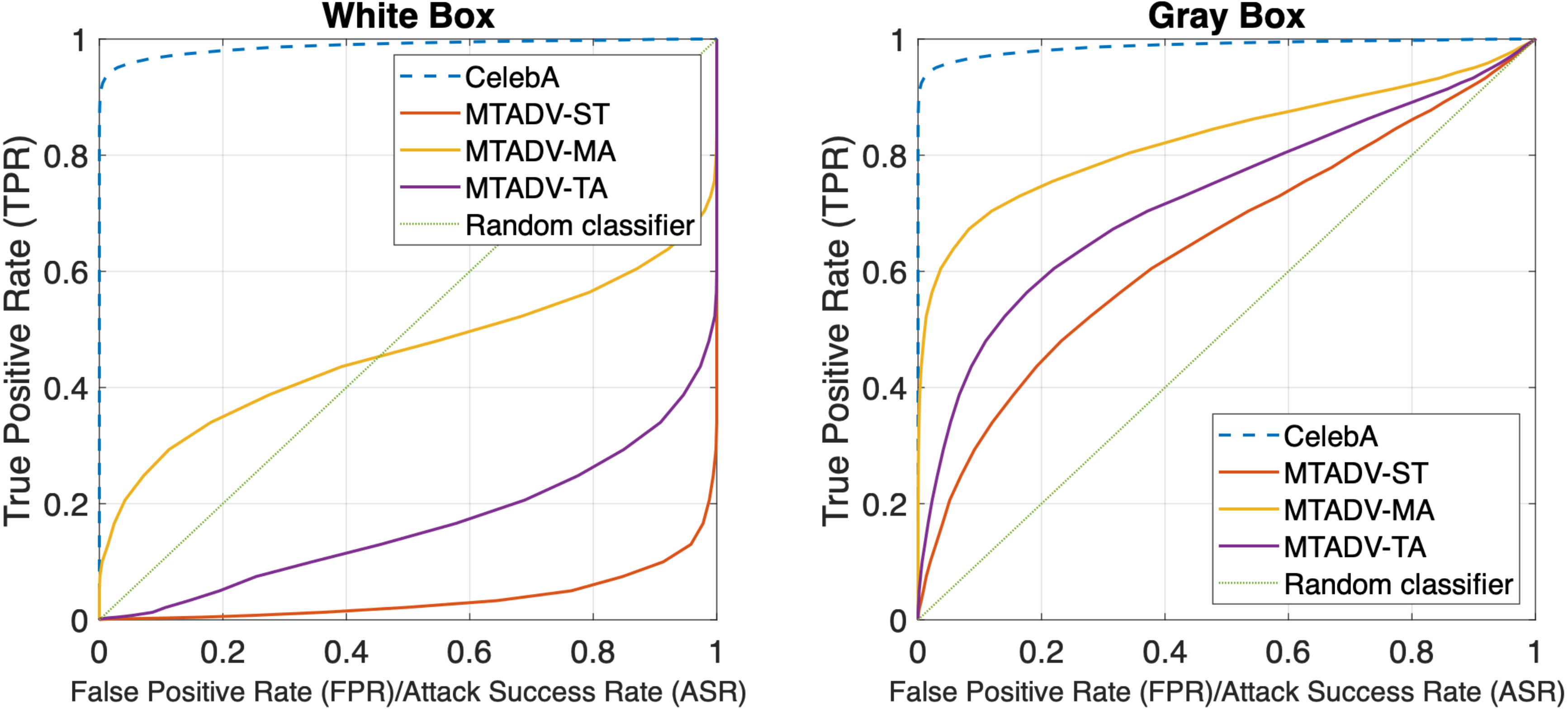}
    \label{fig_ablation_roc_cf}}
    \Description{ROC curves of the CelebA dataset under the proposed attack MTADV}
    \caption{ROC curves of the CelebA dataset under the proposed attack MTADV, compared with that not under attack (blue dashed line).}
    \label{fig_ablation_roc}
\end{figure}

In authentication systems, the threshold is a system setting associated with the security and accuracy. For any adversarial attack, a larger similarity threshold is more likely to protect the system from attacks, but leads to a deterioration of authentication accuracy. The impact of the threshold on the security (ASR or FPR) and accuracy (TPR) can be illustrated by the ROC curves, as shown in Fig. \ref{fig_ablation_roc}. It is observed that in all settings and attack scenarios, it is likely impractical to protect the authentication system from MTADV by adjusting the threshold, as in this case, the TPR dramatically drops simultaneously. It demonstrates the need for a trade-off between security (low attack success rate) and performance (high authentication accuracy).

\subsection{Discussion of countermeasures}
\label{defence}
In Section \ref{performance_asr}, MTADV has been demonstrated applicable as a counterattack to defeat adversarial defenses. In this section, we further review other countermeasures against adversarial attacks. Notably, we justify whether the proposed MTADV can plausibly evade more existing countermeasures.

A well-known class of countermeasures is to introduce an auxiliary model that implements two strategies. 1) Adopt other deep learning models. Hence, a randomly selected deep learning model can be used \cite{dhillon2018stochastic}. Under this strategy, the adversarial examples generated based on one model are unlikely to pass through when the other model is selected. 2) Introduce an adversarial example detection model. Hence, the adversarial examples can be rejected before the next step \cite{Gro17}. However, these countermeasures are unlikely to be feasible for the proposed MTADV as MTADV can attack multiple models simultaneously.

Another countermeasure approach is to detect adversarial examples based on statistical analysis \cite{Gro17}. Specifically, this approach distinguishes adversarial examples from genuine images by observing the inconsistency in terms of distributions or accuracy. However, the experimental results in Table \ref{tab_ST_compare} show that the adversarial examples generated by MTADV are nearly indistinguishable from the genuine target image in the feature space (referring to the column of dissimilarity), and the discrepancy between the adversarial examples and source images is negligible (referring to the columns of SSIM and LPIPS). Thus, identifying the inconsistency between genuine images and adversarial examples generated by MTADV is hardly feasible.

The most plausible countermeasure against MTADV is the adversarial purification that preprocesses the inputs (both genuine images and adversarial examples) to erase the perturbations. Various preprocessing techniques have been reported, for instance, a non-smooth or non-differentiable transformation \cite{Buc18}, a replacement of the input \cite{samangouei2018defense}, a regular or irregularly nullifying \cite{Wan17,XuW171}, or a distortion of the images \cite{LiX17}. However, these methods inevitably cause accuracy deterioration as a price of the security gain.

Another class of countermeasures is to increase the robustness of deep learning models, which attempt to recognize the adversarial examples effectively. One way to increase the robustness of deep learning models is to optimize the architecture of the deep learning models against adversarial attacks, which is rather challenging. Alternatively, a more practical solution is to train the adversarial examples at the training stage \cite{madry2018towards,wong2020fast}. Theoretically, this class of countermeasures might be feasible to resist MTADV. However, the training requires a great number of adversarial examples with sufficient diversity to avoid catastrophic overfitting \cite{wong2020fast}.
\section{Conclusion}
\label{Conclusion}
In this paper, we firstly interpret an overlooked family of impersonation attacks as multi-task adversarial attacks, whose scenarios existing adversarial attacks are inapplicable to. To fill this gap, we propose an adversarial attack, i.e., MTADV, and a novel objective function, which utilises various multi-task attack scenarios by a single algorithm. We further investigate five representative attack scenarios by MTADV, i.e., single-task impersonation attack, morphing attack, universal attack, transferable attack, and counterattack, and the results demonstrate that MTADV is feasible to all these scenarios and general against various datasets and deep learning models. We expect our proposed method can serve as an evaluation tool to assess the security and robustness of a face authentication system. Last but not least, this work further highlights the pitfalls and limitations of existing deep learning models. The extensive analysis performed to examine the vulnerabilities of these models against adversarial attacks could facilitate the development of more robust ones.

\begin{acks}
This work was partially supported by:
\begin{itemize}
    \item The JSPS KAKENHI Grant JP21H04907, and by JST CREST Grants JPMJCR18A6 and JPMJCR20D3, Japan.
    \item The National Natural Science Foundation of China (Nos. 62376003) and Anhui Provincial Natural Science Foundation (No. 2308085MF200).
    \item The Faculty Initiatives Research, Monash University, via Contract No. 2901912.
    \item The framework of the project e.INS-Ecosystem of Innovation for Next Generation Sardinia (cod. ECS 00000038) funded by the Italian Ministry for Research and Education (MUR) under the National Recovery and Resilience Plan (NRRP)-MISSION 4 COMPONENT 2, ``From research to business" INVESTMENT 1.5, ``Creation and strengthening of Ecosystems of innovation", and construction of ``Territorial R\&D Leaders."
\end{itemize}
\end{acks}

\bibliographystyle{ACM-Reference-Format}
\bibliography{sample-base}


\end{document}